\documentclass[lettersize,journal]{IEEEtran}
\usepackage{amsmath,amsfonts,bm}
\usepackage{algorithmic}
\usepackage{algorithm}
\usepackage{array}
\usepackage{textcomp}
\usepackage{stfloats}
\usepackage{url}
\usepackage{verbatim}
\usepackage{graphicx}
\usepackage{cite}
\usepackage[capitalise]{cleveref}
\crefname{assumption}{Assumption}{Assumptions}
\crefname{problem}{Problem}{Problems}
\usepackage{selinput}\SelectInputMappings{adieresis={ä},germandbls={ß}}
\usepackage{epsfig}
\usepackage{times}
\usepackage{float}
\usepackage{tikz,pgfplots}
\usepackage[english]{babel}
\usepackage{verbatim}
\usepackage{makecell}
\usepackage{multirow}
\usepackage{xcolor}
\usepackage{enumitem}
\usepackage{wrapfig}
\usepackage{caption}
\usepackage{subcaption}
\usetikzlibrary{arrows,shapes,backgrounds,patterns,fadings,matrix,arrows,calc,
	intersections,decorations.markings,
	positioning,arrows.meta}
\usepgfplotslibrary{fillbetween}
\usepgfplotslibrary{statistics}
\pgfplotsset{width=5\columnwidth /5, compat = 1.13,
	height = 60\columnwidth /100, grid= major,
	legend cell align = left, ticklabel style = {font=\scriptsize},
	every axis label/.append style={font=\small},
	legend style = {font={\scriptsize}},title style={yshift=-7pt, font = \small} }

\newtheorem{assumption}{\bf{Assumption}}
\newtheorem{theorem}{\bf{Theorem}}

\newtheorem{lemma}{\bf{Lemma}}
\newtheorem{remark}{\bf{Remark}}
\newtheorem{definition}{\bf{Definition}}
\newtheorem{proposition}{\bf{Proposition}}

\newtheorem{problem}{\bf{Problem}}

\crefname@preamble{assumption}{Assumption}{Assumptions}
\crefname@preamble{lemma}{Lemma}{Lemmas}
\crefname@preamble{property}{Property}{Properties}

\DeclareMathOperator*{\argmin}{arg\,min}

\begin{document}
	
	\title{SafeFlow: Safe Robot Motion Planning with Flow Matching via Control Barrier Functions}
	
	\author{
		Xiaobing Dai$^{*}$, 
		Zewen Yang$^{*\dagger}$,~\IEEEmembership{Member,~IEEE,}
		Dian Yu$^{*}$,
		Fangzhou Liu,~\IEEEmembership{Member,~IEEE,} \\
		Hamid Sadeghian,
		Sami Haddadin,~\IEEEmembership{Fellow,~IEEE,}
		Sandra Hirche,~\IEEEmembership{Fellow,~IEEE}
		\thanks{
			$^{*}$Equal contribution. $^{\dagger}$Corresponding author \textless  zewen.yang@tum.de \textgreater.
		}
		\thanks{
			Xiaobing Dai and Sandra Hirche are with the Chair of Information-oriented Control (ITR), School of Computation, Information and Technology (CIT), Technical University of Munich (TUM), 80333 Munich, Germany.
		}
		\thanks{
			Zewen Yang and Hamid Sadeghian are with the Chair of Robotics and Systems Intelligence, Munich Institute of Robotics and Machine Intelligence (MIRMI), Technical University of Munich (TUM), 80992 Munich, Germany.
	}
		\thanks{
	Dian Yu is with the Professorship of AI Planning in Dynamic Environments, School of Computation, Information and Technology (CIT), Munich Institute of Robotics and Machine Intelligence (MIRMI),  Technical University of Munich (TUM), 80992 Munich, Germany.
			}
			\thanks{
			Fangzhou Liu is with the National Key Laboratory of Modeling and Simulation for Complex Systems, Harbin Institute of Technology, 150001, P.R.China.
		}
	\thanks{
			Sami Haddadin is with the Mohamed Bin Zayed University of Artificial Intelligence, Abu Dhabi 23201, United Arab Emirates.
		}
	}

	\maketitle
	
	\begin{abstract}
		Recent advances in generative modeling have led to promising results in robot motion planning, particularly through diffusion and flow matching (FM)-based models that capture complex, multimodal trajectory distributions. 
		However, these methods are typically trained offline and remain limited when faced with new environments with constraints, often lacking explicit mechanisms to ensure safety during deployment. 
		In this work, safe flow matching (SafeFlow), a motion planning framework, is proposed for trajectory generation that integrates flow matching with safety guarantees. 
		SafeFlow leverages our proposed flow matching barrier functions (FMBF) to ensure the planned trajectories remain within safe regions across the entire planning horizon. 
		Crucially, our approach enables training-free, real-time safety enforcement at test time, eliminating the need for retraining.
		We evaluate SafeFlow on a diverse set of tasks, including planar robot navigation and 7-DoF manipulation, demonstrating superior safety and planning performance compared to state-of-the-art generative planners. 
		Comprehensive resources are available on the project website: \url{https://safeflowmatching.github.io}.
	\end{abstract}
	
	
	\section{Introduction}
	
	\IEEEPARstart{E}{nabling} robots to autonomously generate safe and feasible motions in dynamic environments remains one of the central challenges in robotic learning and motion planning. 
	Recent advances in deep generative modeling, particularly in diffusion models, have significantly expanded the capabilities of robots to learn diverse and high-dimensional behaviors from demonstrations or interactions. 
	
	As a commonly used generative model, diffusion models have proven effective in modeling multimodal trajectory distributions, enabling robots to adaptively generate context-dependent motions across various tasks, such as navigation~\cite{Sridhar2024ICRANoMaD}, object manipulation~\cite{Janner2022ICMLplanning}, and locomotion~\cite{gadginmath2025dynamicsawarediffusionmodelsplanning}. 
	Despite their success, diffusion models come with notable limitations. 
	Specifically, the inference process is typically slow and computationally intensive, as it involves iteratively solving a stochastic differential equation over many denoising steps~\cite{song2021ICLRscorebased}. 
	Furthermore, while diffusion models can be conditioned on goals or task descriptions, they generally lack formal safety guarantees, which is crucial when deploying robots in dynamic environments with obstacles, humans, or other agents. 
	Recent works have begun to address these concerns by incorporating safety constraints into diffusion-based frameworks (e.g., via model predictive control~\cite{zhou2024diffusionmodelpredictivecontrol} or barrier functions~\cite{Glotfelter2017CSLnonsmooth}), but such repeated inference using diffusion models further increases computational complexity and may compromise inference quality or task performance.
	
\begin{figure}[t]
		\centering
		\includegraphics[width=0.45\textwidth]{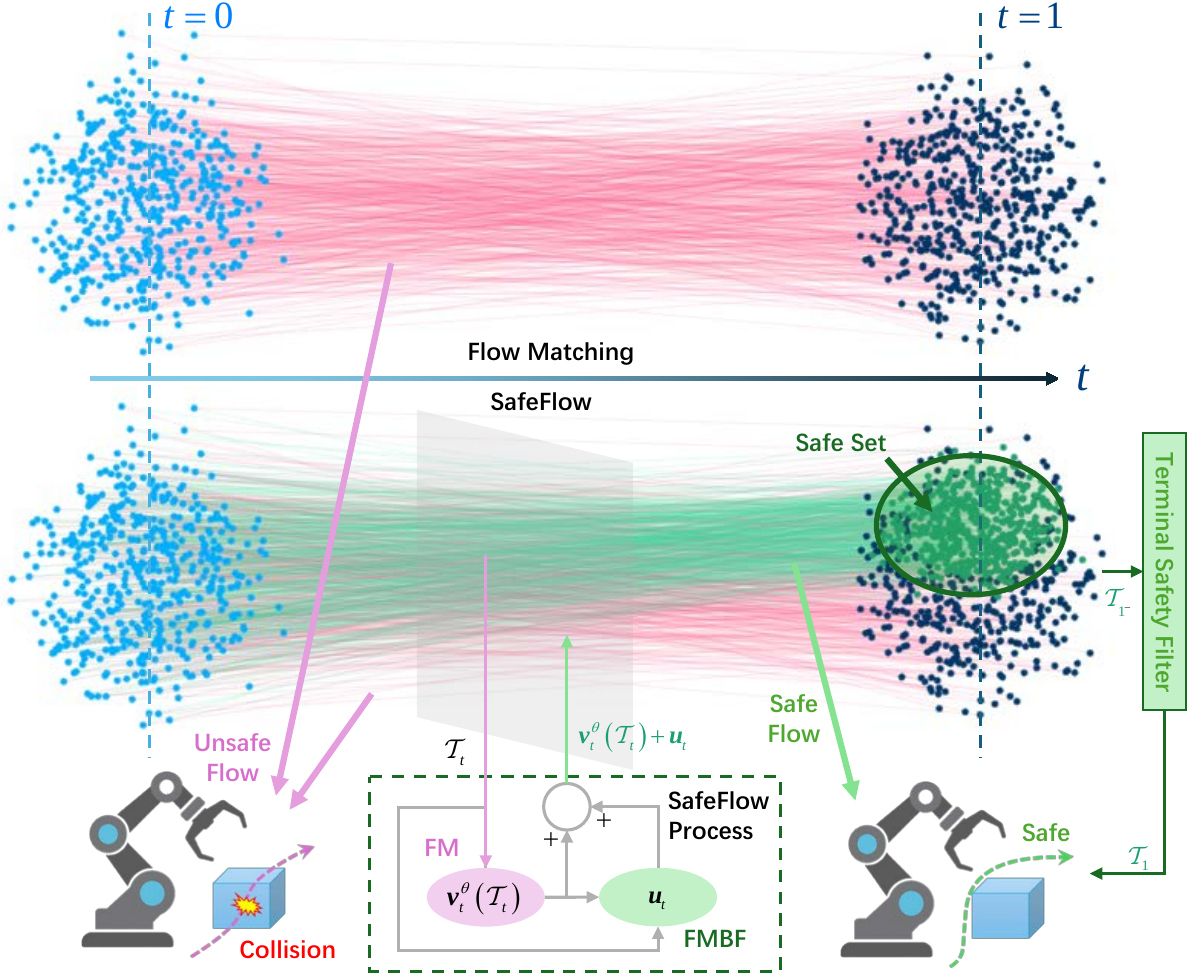}
		\caption{Illustration of SafeFlow process.}
		\vspace{-0.3cm}
		\label{fig_intro}
	\end{figure}

	To overcome these challenges, we turn to flow matching (FM), which is a recently emerging generative modeling framework offering a promising alternative to diffusion models~\cite{lipman2023ICLRflow}. 
	Instead of relying on stochastic processes, FM defines a deterministic flow field—governed by an ordinary differential equation—that smoothly transforms samples from a simple prior distribution into target data samples~\cite{Kornilov2024NeurIPSOptimal}. 
	More importantly, the FM framework is amenable to explicit constraint integration, making it a natural candidate for generating safe robot motions~\cite{liu2023ICLRflow, Xie2024ICMLReflected} as shown in \cref{fig_intro}. 
	
	\subsection{Related Work}
	Generative models for robot learning have gained considerable attention in recent years. 
	This section reviews recent advancements in diffusion-based and flow-based methods, with a focus on their applications to robot learning.

	\subsubsection{Diffusion-based Models for Robot Learning.} 
	Diffusion models have been effectively utilized in various robot learning domains, including motion planning, imitation learning, and reinforcement learning, as demonstrated by works such as~\cite{Janner2022ICMLplanning, gadginmath2025dynamicsawarediffusionmodelsplanning, Carvalho2023IROSMotion}.
	Especially in imitation learning, diffusion models have been leveraged to produce robot trajectories based on expert demonstrations. 
	For instance, diffusion models are employed in \cite{pearce2023ICLRimitating, reuss2023RSSgoal} to develop policies that replicate expert behaviors.
	For robotic manipulation tasks, ChainedDiffuser is introduced in \cite{zhou2023CoRLchainedDiffuser}, which is a novel policy architecture integrating action keypose prediction with trajectory diffusion generation. 
	To tackle long-horizon planning and task execution, skill-centric diffusion models are proposed in \cite{li2023ICMLhierarchical, Mishra2023CoRLGenerative}, that combine learned distributions to create extended plans during inference. 
	For generating control actions, diffusion policy is presented in \cite{chi2024IJRRdifusion}, which frames policy learning as a conditional denoising diffusion process over the action space of the robot, conditioned on 2D observation features. 
	This concept is advanced by \cite{Ze2024RSSDP3} in their 3D Diffusion Policy, which incorporates complex 3D visual representations to enable generalizable visuomotor policies from just a few dozen expert demonstrations.
	However, these methods ignore the nature of system dynamics and safety requirement in practical operation environment.
	
	Recent efforts have also aimed to ensure that generated trajectories comply with the physical laws governing robot dynamics. 
	For instance, system dynamics is embedded directly into the denoising process as shown in \cite{gadginmath2025dynamicsawarediffusionmodelsplanning}, while generated state sequences are used in \cite{chen2025ICRAlearningcoordinatedbimanualmanipulation}with an inverse dynamics model to derive actions for coordinated bimanual tasks.
	To enforce safety and dynamical constraints, penalties for constraint violations are incorporated into the loss function \cite{li2024constraintawarediffusionmodelstrajectory}, though this method does not guarantee strict adherence to hard constraints. 
	Alternatively, generated policies are projected onto a dynamically admissible manifold during training and inference \cite{bouvier2025ddatdiffusionpoliciesenforcing}. 
	However, this technique under-approximates the reachable set using a polytope, relying on the assumption of smooth nonlinear dynamics and environments. 
	Drawing inspiration from CBFs, SafeDiffuser and CoBL are proposed in \cite{xiao2025ICLRsafediffuser, Mizuta2024IROScoBLDiffusion} respectively, which are aimed at enforcing safety constraints within reverse diffusion models for safe planning. 
	However, a critical technical limitation exists.
	Specifically, CBFs rely on the assumption of initially safe samples, which is fundamentally violated for the reverse diffusion process where the initial sample is usually obtained from unbounded Gaussian distribution.
	Furthermore, due to the probabilistic nature of diffusion models, there is no guarantee that the generated trajectories will remain within the safe set throughout the process.
	As a result, the practical applications of diffusion model-based framework to real-world safe robot learning and control remain limited.

	\subsubsection{Flow-based Models for Robot Learning} 
	
	For deterministic generation performance, flow matching models have emerged as a promising alternative to diffusion models, offering potential improvements in efficiency and performance for robotic applications.
	For robotic applications, flow matching policy is introduced in \cite{zhang2025affordancebasedrobotmanipulationflow}, that utilizes affordance-based reasoning for manipulation tasks. 
	Moreover, departing from image-based inputs in \cite{zhang2025affordancebasedrobotmanipulationflow}, conditional flow matching is employed with point clouds in \cite{chisari2024CoRLlearning} to generate robot actions. 
	Furthermore, to enhance generation efficiency, a flow policy that uses consistency flow matching with a straight-line flow is proposed in \cite{zhang2024AAAIflowpolicyenablingfastrobust}, enabling one-step generation and striking a balance between speed and effectiveness. 
	However, limitations persist in complex manipulation scenarios.
	To this end, a Riemannian manifold for robot state representation \cite{Braun2024IROSRiemannian} is integrated in flow matching, which better captures high-dimensional multimodal distributions and improves the handling of complex robotic systems.
	While flow matching improves the generation efficiency and the smoothness of the generated results, the integration of the physical property and the safety of robot remains unaddressed.
	
	\subsection{Contribution}
	
	In this work, a novel method, i.e., safe flow matching (SafeFlow), is introduced for constraint-aware motion planning.
	SafeFlow combines the efficiency and flexibility of FM with the formal safety guarantees provided by the proposed flow matching barrier functions (FMBFs), inspired by CBFs.
	Specifically, it learns a continuous-time vector field that maps random waypoints from a prior distribution to goal-directed robot trajectories, while simultaneously ensuring that the generated trajectories respect safety constraints. 
	By leveraging finite-time flow invariance principles~\cite{lipman2023ICLRflow}, a regularization term is embedded directly into the FM model, enabling real-time generation of safe and feasible motions without sacrificing generation quality or expressiveness.
	We compare SafeFlow to state-of-the-art (SOTA) methods, demonstrating the superiority of our method across six performance metrics, e.g., safety, smoothness, and efficiency. 
	The main contributions of this paper are:
	\begin{itemize}[left=4pt,itemsep=0pt]
		\item SafeFlow, a novel FM framework for robot motion planning under safety constraints.
		\item A theoretically grounded integration of control barrier functions, as a plug-and-play safety module, into flow-based generative models to ensure safe trajectory generation.
		\item Empirical demonstrations on the superior safety capabilities and planning performance of the proposed SafeFlow across multiple tasks.
	\end{itemize}
	
	\section{Preliminaries}
	\subsection{Flow Matching Model}
	In this paper, a robot is considered with states $\bm{s} \in \mathbb{S} \subseteq \mathbb{R}^{d_s}$, where $d_s \in \mathbb{N}_{>0}$ denotes the state dimension.
	It is intended to generate the robot motion trajectory $\bm{\mathcal{T}} \in \mathbb{S}^{H+1} \subseteq \mathbb{R}^d$.
	The trajectory $\bm{\mathcal{T}}$ is represented as a state sequence with  a horizon $H \in \mathbb{N}_{>0}$, i.e., $\bm{\mathcal{T}} = [(\bm{s}^0)^\top, \dots, (\bm{s}^H)^\top]^{\top}$ with $d = (H + 1) d_s$, where $\bm{s}^k \in \mathbb{S}$ is the state at the $k$-th step with $k = 0, \cdots, H$.
	Following the flow matching model, the trajectory $\bm{\mathcal{T}}$ is generated by solving an ordinary differential equation (ODE) governed by a time-dependent vector field $\bm{v}(\cdot, \cdot): [0, 1] \times \mathbb{R}^d \to \mathbb{R}^d$, which induces a flow map $\bm{\psi}(\cdot, \cdot): [0, 1] \times \mathbb{R}^d \to \mathbb{R}^d$ satisfying $\bm{\psi}(0, \bm{\mathcal{T}}) = \bm{\mathcal{T}}$ for all $\bm{\mathcal{T}} \in \mathbb{R}^d$.
	Specifically, the flow matching model is written as
	\begin{align} \label{eqn_FM}
		\mathrm{d} \bm{\psi}(t, \bm{\mathcal{T}}_0) / \mathrm{d} t = \bm{v}(t, \bm{\psi}(t, \bm{\mathcal{T}}_0)), && \forall t \in [0,1],
	\end{align}
	where $\bm{\mathcal{T}}_0$ is sampled from a known prior distribution $p$, i.e., $\bm{\mathcal{T}}_0 \sim p$.
	For notational simplicity, denote $\bm{\mathcal{T}}_t = \bm{\psi}(t, \bm{\mathcal{T}}_0)$, which is also written as $\bm{\mathcal{T}}_t = [(\bm{s}_t^0)^\top, \dots, (\bm{s}_t^H)^\top]^{\top}$ with $\bm{s}_t^k \in \mathbb{R}^{d_s}$ for all $t \in [0,1]$ and $k = 0, \cdots, H$.
	Then, the generated trajectory is regarded as $\bm{\mathcal{T}} = \bm{\mathcal{T}}_1$.
	Note that the exact expression of the flow $\bm{v}(\cdot,\cdot)$ is hard to obtain, but an approximated model $\bm{v}^{\bm{\theta}}(\cdot,\cdot)$ with parameters $\bm{\theta} \in \mathbb{R}^{d_{\theta}}$ and $d_{\theta} \in \mathbb{N}_{>0}$ can be learned using a neural network \cite{lipman2023ICLRflow} from the data set $\mathbb{D}$ collecting expert demonstrations, i.e., $\mathbb{D} = \{ \bm{\mathcal{T}}_0^{(\iota)}, \bm{\mathcal{T}}^{(\iota)} \}_{\iota \in \mathbb{N}}$.
	The collected demonstrations $\bm{\mathcal{T}}^{(\iota)}$ are considered to follow a target unknown distribution $q$, and the intermediate trajectories $\bm{\mathcal{T}}_t$ form a dsitribution $p_t$ for all $t \in [0,1]$.
	Therefore, the objective of flow matching is to train a parameterized vector field $\bm{v}_t^{\bm{\theta}} = \bm{v}^{\bm{\theta}}(t,\cdot)$ such that its induced flow $\bm{\psi}_t = \bm{\psi}(t, \cdot)$ generates a probability path $p_t$, transitioning from an initial distribution $p_0$ to the target data distribution $q$.
	
	\subsection{Safe Invariance in Dynamical Systems}
	\label{subsection_CBF}
	
	To ensure the safety of the generated trajectory $\bm{\mathcal{T}}$, control barrier functions are introduced in this subsection. 
	In particular, consider a control-affine system
	\begin{align}
		\label{eq_dynamics2}
		\dot{\bm{x}}(t) = \bm{f}(\bm{x}(t)) + \bm{g}(\bm{x}(t)) \bm{u}(t),
	\end{align}
	where $\bm{x}(\cdot): \mathbb{R}_{\ge 0} \to \mathbb{X} \subset \mathbb{R}^{d_x}$ and $\bm{u}(\cdot): \mathbb{R}_{\ge 0} \to \mathbb{R}^{d_u}$ are system states and control inputs with $d_x, d_u \in \mathbb{N}_{>0}$ respectively.
	The functions $\bm{f}(\cdot):\mathbb{X} \to \mathbb{R}^{d_x}$ and $\bm{g}(\cdot): \mathbb{X} \to \mathbb{R}^{d_x \times d_u}$ are assumed to be known and locally Lipschitz. 
	For notational simplicity, denote $\bm{x}_t = \bm{x}(t)$ and $\bm{u}_t = \bm{u}(t)$ for all $t \in \mathbb{R}_{\ge 0}$.
	Moreover, a super level set $\mathbb{C}:=\{\bm{x}\in \mathbb{X} | h(\bm{x})\ge 0 \}$ is defined via a continuously differentiable equation $h(\cdot):\mathbb{X}\!\to\! \mathbb{R}$, which describes the safe state set.
	The classic safety requirement for \eqref{eq_dynamics2} is defined as forward invariance of $\mathbb{C}$, which is ensured using CBF-based method defined as follows.
	
	\begin{definition}[Control Barrier Function \cite{Glotfelter2017CSLnonsmooth}]
		\label{def_cbf}
		Given a safe set $\mathbb{C}$, $h(\cdot)$ is a time-invariant CBF for \eqref{eq_dynamics2}, if there exists a class $\mathcal{K}$ function $\alpha(\cdot)$ such that for all $t \in \mathbb{R}_{\ge 0}$ it has
		\begin{align} \label{eq_cbf}
			\sup\nolimits_{\bm{u} \in \mathbb{R}^m} \{ L_{\bm{f}} h(\bm{x}_t) + L_{\bm{g}} h(\bm{x}_t) \bm{u} + \alpha(h(\bm{x}_t)) \} \ge 0,
		\end{align}
		where $L_{\bm{f}} h(\bm{x}_t), L_{\bm{g}} h(\bm{x}_t)$ denote the Lie derivatives of $h(\bm{x}_t)$ along $\bm{f}(\bm{x}_t)$ and $\bm{g}(\bm{x}_t)$, respectively.
	\end{definition}
	
	With the CBF condition \eqref{eq_cbf} shown in \cref{def_cbf}, the safety of the system \eqref{eq_dynamics2} is derived in the following lemma.
	
	\begin{lemma} \label{lemma_CBF}
		The system \eqref{eq_dynamics2} is safe with forward invariant $\mathbb{C}$, if $\bm{x}(0) \in \mathbb{C}$ and $h(\cdot)$ is a control barrier function.
	\end{lemma}
	
	\cref{lemma_CBF} shows the relationship between CBF and the system safety.
	Note that the initial condition $\bm{x}(0) \in \mathbb{C}$ is essential to maintain the system safety, which is not naturally satisfied in flow matching-based generation task.
	
	\subsection{Trajectory Generation Objective}
	
	In this work, we aim to generate a trajectory $\bm{\mathcal{T}}$ that closely aligns with a target distribution $q$ while remaining within a predefined safe set.
	Specifically, we first define the safety of a trajectory $\bm{\mathcal{T}}$.
	Consider a safe set $\mathbb{C}_s \subseteq \mathbb{S}$ defined for each state $\bm{s}^i$, $i = 0, \dots, H$, which is characterized by a single continuous differentiable function $h(\cdot): \mathbb{S} \to \mathbb{R}$, i.e., 
	\begin{align}
		\mathbb{C}_s = \{ \bm{s} \in \mathbb{S} | h(\bm{s}) \ge 0 \}.
	\end{align}
	The safe set $\mathbb{C}_s$ satisfies the following assumption.
	
	\begin{assumption} \label{assumption_safe_set}
		$\mathbb{C}_s$ is non-empty and has no isolated points.
	\end{assumption}
	
	\cref{assumption_safe_set} is commonly found in most safe control problem \cite{zhang2025gaussian, huang2024learning}, where the non-empty property ensures the feasibility of safety constraint.
	The exclusion of the isolated points guarantees that the system can move continuously within $\mathbb{C}_s$.
	Therefore, \cref{assumption_safe_set} is not restrictive.
	
	The generated trajectory $\bm{\mathcal{T}}$ is called safe if all generated states $\bm{s}^i$ remain safe w.r.t $\mathbb{C}_s$ and $h(\cdot)$, which is mathematically defined as follows.
	\begin{definition}[Safe Trajectory and Safe Flow Matching] 
		\label{definition_safe_flow_matching}
		A trajectory $\bm{\mathcal{T}}$ with a horizon $H+1$ is safe if
		\begin{align}
			h(\bm{E}_k \bm{\mathcal{T}}) \geq 0, && \forall k = 0, \cdots, H,
		\end{align}
		where $\bm{E}_k \in \mathbb{R}^{d_s \times d}$ are selection matrices defined as $\bm{E}_k = [\bm{0}_{d_s \times k d_s}, \bm{I}_{d_s}, \bm{0}_{d_s \times (H-k) d_s}]$ such that $\bm{s}^k = \bm{E}_k \bm{\mathcal{T}}$.
	\end{definition}
	
	The objective is to propose a safe flow matching framework for robot motion generation, which is defined as follows.
	
	\begin{problem}
		Design a flow matching framework for robot motion planning, such that all the generated trajectory $\bm{\mathcal{T}} = \bm{\mathcal{T}}_1$ at $t = 1$ are safe for any initial sample $\bm{\mathcal{T}}_0 \sim p_0$.
		Such generation process is called as safe flow matching.
	\end{problem}
	
	\section{Safe Flow Matching}
	
	In this section, a safe flow matching framework is proposed. 
	First, the concept of flow matching barrier functions (FMBFs) is introduced, followed by the theoretical guarantee of safe generation. 
	Furthermore, safety regularization terms are designed to enforce compliance with safety constraints during trajectory generation. 
	\subsection{Safe Flow Matching under Unified Constraints}
	
	It is easy to see that the flow matching model \eqref{eqn_FM} is not naturally safe as in \cref{definition_safe_flow_matching} due to its training process and potentially unobserved obstacles in data set.
	To ensure that all trajectories $\bm{\mathcal{T}}$ generated from FM are safe, we draw inspiration from the safety guarantees provided by CBF-based methods and propose an adaptation for trajectory generation. 
	Specifically, a regularization term $\bm{u}_t = [{\bm{u}_t^0}^{\top}, \cdots, {\bm{u}_t^H}^{\top}]^{\top} \in \mathbb{R}^d$ with $\bm{u}_t^k \in \mathbb{R}_{d_s}$ and $k = 0, \cdots, H$ is introduced at any generation step $t \in [0,1]$.
	The introduced $\bm{u}_t$ steers the flow toward a safe manifold, such that the guided flow matching model is formulated as
	\begin{align}
		\label{eq_controlled_dynamics}
		\mathrm{d} \bm{\psi}_t( \bm{\mathcal{T}}_0) / \mathrm{d} t = \bm{v}_t^{\bm{\theta}}(\bm{\psi}_t( \bm{\mathcal{T}}_0)) + \bm{u}_t, &&\forall t \in [0,1]
	\end{align}
	with $\bm{\mathcal{T}}_0 \sim p_0$.
	To ensure the safety, the guidance term $\bm{u}_t$ is obtained by using CBF-based methods in \cref{lemma_CBF}.
	Notably, the initial safety condition $\bm{x}_0 \in \mathbb{C}$ is generally not satisfied in particularly in the reverse diffusion process considering $\bm{x}_t = \bm{\mathcal{T}}_t$, where the initial $\bm{\mathcal{T}}(\cdot)_0$ is randomly sampled from an unbounded distribution $p_0$.
	This initial sample insecurity has been overlooked in previous works~\cite{xiao2025ICLRsafediffuser, Mizuta2024IROScoBLDiffusion}.
	Thankfully, the task of driving the system \eqref{eq_controlled_dynamics} into the safe set during the reverse diffusion process can be defined as prescribed-time safety as follows.
	
	\begin{definition}[Prescribed-time Safety \cite{Abel2024TACPrescribed}]
		\label{definition_PT_Safety}
		A system \eqref{eq_dynamics2} with arbitrary initial state $\bm{x}(0) \in \mathbb{X}$ is prescribed-time safe with a predefined time $T_d \in \mathbb{R}_{>0}$, if $\bm{x}_t \in \mathbb{C}$ for any $t \ge T_d$.
	\end{definition}
	
	The condition for prescribed-time safety in \cref{definition_PT_Safety} is divided into two parts: entering the safe set $\mathbb{C}$ before $T_d$ and maintaining in $\mathbb{C}$ after the entrance.
	While the second part can be achieved using CBF defined in \cref{def_cbf}, the prescribed-time entrance into the safe set $\mathbb{C}$ from an unsafe initial state requires modification on the CBF condition \eqref{eq_cbf}.	
	Specifically, we propose an FMBF to enforce safety constraints in the flow matching framework, which is defined as follows.
	
	\begin{definition} [Flow Matching Barrier Function] \label{definition_flow_matching_barrier_function}
		A function $h(\cdot)$ is called a flow matching barrier function, if 
		\begin{align}
			\sup_{\bm{u}_t^k \in \mathbb{R}^{d_s}} \Big\{ \frac{\mathrm{d} h(\bm{E}_k \bm{\mathcal{T}}_t)}{\mathrm{d} \bm{E}_k \bm{\mathcal{T}}_t} &\bm{E}_k \bm{v}_t^{\bm{\theta}}(\bm{\mathcal{T}_0}) + \frac{\mathrm{d} h(\bm{E}_k \bm{\mathcal{T}}_t)}{\mathrm{d} \bm{E}_k \bm{\mathcal{T}}_t} \bm{u}_t^k \\
			&+ \varphi(t, h(\bm{E}_k \bm{\mathcal{T}}_t)) h(\bm{E}_k \bm{\mathcal{T}}_t) \Big\} \ge 0, \nonumber
		\end{align}
		hold for all $k = 0, \cdots, H$ and $t \in [0,1)$.
		The function $\varphi(\cdot, \cdot): [0,1) \times \mathbb{R} \to \mathbb{R}_{> 0}$ is chosen as
		\begin{align} \label{eqn_varphi}
			\varphi(t, h) = \begin{cases}
				\varphi_0, & \text{if}~ h \ge 0 \\
				\varphi_1(t), & \text{otherwise}
			\end{cases}, && \forall t \in [0,1),
		\end{align}
		where $\varphi_0 \in \mathbb{R}_{>0}$ and $\varphi_1(\cdot): [0,1) \to \mathbb{R}_{>0}$ are chosen as monotonically increasing and satisfying $\int_0^{1^-} \varphi_1(s) \mathrm{d} s \!=\! \infty$.
	\end{definition}
	
	Unlike the class $\mathcal{K}$ function used in CBF condition \eqref{eq_cbf} in \cref{def_cbf}, the function $\varphi(\cdot, \cdot)$ in \eqref{eqn_varphi} is more complicated due to the allowable negative $h(\cdot)$.
	For safe state $\bm{s} \in \mathbb{C}_s$, the function $\varphi(t, h(\bm{s})) h(\bm{s}) = \varphi_0 h(\bm{s})$ belongs to class $\mathcal{K}$ w.r.t $h(\bm{s})$ for all $t \in [0,1)$.
	In the case with $\bm{s} \notin \mathbb{C}_s$ resulting in $h(\bm{s}) < 0$, the function $\varphi(\cdot, h(\bm{s})) = \varphi_1(\cdot)$ behaves as a blow-up function, whose value goes to infinity when $t \to 1^-$, i.e., $\lim_{t \!\to\! 1^-} \varphi_1(t) \!=\! \infty$.
	In recent works, some examples of $\varphi_1(\cdot)$ are shown, e.g., in exponential form \cite{zhang2025gaussian} as
	\begin{align} \label{eqn_varphi_1_exponential}
		\varphi_1(t) = - \dot{\xi}(t) / \xi(t) && \text{with} &&\xi(t) = \exp(\omega (1 - t)) - 1
	\end{align}
	and in inverse polynomial form \cite{huang2024learning} as
	\begin{align} \label{eqn_varphi_1_polynomial}
		\varphi_1(t) = \omega / (1 - t)^2
	\end{align}
	with $\omega \in \mathbb{R}_{> 0}$.
	Building upon the FMBF, the safety guarantee of the generated trajectory is derived in the following theorem.
	
	\begin{theorem}
		\label{theorem_FMBF}
		Suppose $h(\cdot)$ is a valid FMBF as \cref{definition_flow_matching_barrier_function} for guided FM \eqref{eq_controlled_dynamics}.
		Choose $\bm{u}_t$ from $\mathbb{U}_t = \prod_{k = 0}^H \mathbb{U}_t^k$ with
		\begin{align}
			\mathbb{U}_t^k = \big\{ \bm{u}_t^k \in \mathbb{R}^{d_s} \Big|& \frac{\mathrm{d} h(\bm{E}_k \bm{\mathcal{T}}_t)}{\mathrm{d} \bm{E}_k \bm{\mathcal{T}}_t} \bm{E}_k \bm{v}_t^{\bm{\theta}}(\bm{\mathcal{T}_0}) + \frac{\mathrm{d} h(\bm{E}_k \bm{\mathcal{T}}_t)}{\mathrm{d} \bm{E}_k \bm{\mathcal{T}}_t} \bm{u}_t^k \nonumber\\
			&+ \varphi(t, h(\bm{E}_k \bm{\mathcal{T}}_t)) h(\bm{E}_k \bm{\mathcal{T}}_t) \ge 0 \big\} 
		\end{align}
		for all $k = 0, \cdots, H$ and $t \in [0,1)$.
		Then, the FM process \eqref{eq_controlled_dynamics} is safe under \cref{assumption_safe_set}, ensuring that all generated trajectories $\bm{\mathcal{T}}_1$ fulfill the safety condition in \cref{definition_safe_flow_matching}.
	\end{theorem}
	\begin{IEEEproof}
		See Appendix A.
	\end{IEEEproof}
	
	\cref{theorem_FMBF} shows the guided FM \eqref{eq_controlled_dynamics} is a safe flow matching, if $\bm{u}_t$ is chosen from $\mathbb{U}_t$, where the non-empty property of $\mathbb{U}_t$ is guaranteed by valid FMBF.
	Although valid FMBFs ensure safety for both the trajectory and the FM process, the selection of an appropriate regularization term $\bm{u}_t$ for each $t \in [0, 1)$ remains a critical challenge. 
	To minimize the influence of $\bm{u}_t$ on the distribution’s alignment with the target $p_1$, the regularization term should be kept as small as possible in magnitude. 
	To this end, we determine $\bm{u}_t$ by solving the following optimization problem
	\begin{align} \label{eqn_optimization}
		&\bm{u}_t = \argmin\nolimits_{\bm{u} = [(\bm{u}^0)^{\top}, \cdots, (\bm{u}^H)^{\top}]^{\top} \in \mathbb{R}^d} \| \bm{u} \|^2 \\
		\text{s.t.~}& a_t^k + (\bm{b}_t^k)^{\top} \bm{u}^k \ge 0, \forall i \in \{ 0, \cdots, H \}, \nonumber
	\end{align}
	with $a_t^k \in \mathbb{R}$ and $\bm{b}_t^k \in \mathbb{R}^{d_s}$ as
	\begin{align}
		&\bm{b}_t^k = (\partial h(\bm{E}_k \bm{\mathcal{T}}_t) / \partial (\bm{E}_k \bm{\mathcal{T}}_t) )^{\top}, \\
		&a_t^k = (\bm{b}_t^k)^{\top} \bm{E}_k \bm{v}_t^{\bm{\theta}}(\bm{\mathcal{T}}_0) + \varphi(t, h(\bm{E}_k \bm{\mathcal{T}}_t)) h(\bm{E}_k, \bm{\mathcal{T}}_t),
	\end{align}
	which can be considered as a quadratic programming (QP) problem due to the quadratic objective function and linear constraint w.r.t $\bm{u}_t$.
	The feasibility of \eqref{eqn_optimization} is shown as follows.
	
	\begin{proposition}
		Suppose that the derivative of $h(\cdot)$ is non-zero and \cref{assumption_safe_set} is satisfied, then $h(\cdot)$ is a valid FMBF as in \cref{definition_flow_matching_barrier_function} for \eqref{eq_controlled_dynamics}.
		Moreover, the QP problem is always feasible with closed-form solution of $\bm{u}_t$ for any $t \in [0,1)$.
	\end{proposition}
	\begin{IEEEproof}
		Given that $\| \bm{u} \|^2 = \sum_{k = 0}^H \| \bm{u}^k \|^2$, the optimization problem can be decoupled for each $\bm{u}^k$ as
		\begin{align} \label{eqn_decoupled_optimization}
			&\bm{u}_t^k = \argmin\nolimits_{\bm{u}^k \in \mathbb{R}^{d_s}} \| \bm{u}^k \|^2 \nonumber \\
			\text{s.t.~}& a_t^k + (\bm{b}_t^k)^{\top} \bm{u}^k \ge 0.
		\end{align}
		This decoupling also indicates the selection of $\bm{u}_t^k$ from $\mathbb{U}_t^k$ individually, i.e., $\bm{u}_t^k \in \mathbb{U}_t^k$ for any $k = 0, \cdots, H$.
		Note that a valid FMBF requires $a_t^k \ge 0$ when $\bm{b}_t^k = \bm{0}_{d_s \times 1}$ for any $k = 0, \cdots, H$ and $t \in [0,1)$.
		Considering $\mathrm{d} h(\bm{s}) / \mathrm{d} \bm{s} \ne \bm{0}_{d_s \times 1}$ for all $\bm{s} \in \mathbb{S}$ and $\mathbb{C} \ne \emptyset$ as shown in the premise, it is easy see $\bm{b}_t^k \ne \bm{0}_{d_s \times 1}$, indicating $h(\cdot)$ is a valid FMBF.
		Then, the solution for such a QP problem \eqref{eqn_decoupled_optimization} always exists and has an explicit form for all $k = 0, \cdots, H$ and $t \in [0,1)$ as
		\begin{align} \label{eqn_u_closed_form}
			\bm{u}_t^k = \begin{cases}
				\bm{0}_{d_s \times 1}, & \text{if~} a_t^k \ge 0 \\
				- {\bm{b}_t^k a_t^k}/{\|{\bm{b}_t^k}\|^2}, & \text{otherwise}
			\end{cases}
		\end{align}
		following Karush-Kuhn-Tucker (KKT) conditions.
		Note that the value of $\varphi(t, \cdot)$ may go to infinity when $t \to 1^-$ due to the blow-up $\varphi_1(\cdot)$, the practical numerical computation methods may suffer instability, which causes computation error and deteriorates trajectory safety, i.e., $\bm{\mathcal{T}}_{1^-} \notin \mathbb{C}_s^{H+1}$.
		To compensate for such potential computation error, a safety filter is designed at the last step for generating an absolutely safe trajectory $\bm{\mathcal{T}} = \bm{\mathcal{T}}_1$.
		Specifically, the generated trajectory is obtained by solving the optimization problem
		\begin{align} \label{eqn_terminal_optimal_problem}
			&\bm{\mathcal{T}}_1 = \argmin\nolimits_{\bm{\mathcal{T}} \in \mathbb{R}^d} \| \bm{\mathcal{T}} - \bm{\mathcal{T}}_{1^-} \| \\
			\text{s.t.~}& h(\bm{E}_k \bm{\mathcal{T}}) \geq 0, \qquad \forall k = 0, \cdots, H,
		\end{align}
		which directly ensure the trajectory safety for $\bm{\mathcal{T}} = \bm{\mathcal{T}}_1$.
		The existence of the solution for \eqref{eqn_terminal_optimal_problem} is ensured due to the non-empty property of $\mathbb{C}$, inducing at least one safe solution as $\bm{\mathcal{T}}_1 = [\bm{s}_{safe}^{\top}, \cdots, \bm{s}_{safe}]^{\top}$ with $\bm{s}_{safe} \in \mathbb{C}$.
		This observation leads to the validity of $h(\cdot)$ as a FMBF.
	\end{IEEEproof}
	
	The entire process of the proposed SafeFlow, including FMBF and terminal safety filter, is shown in \cref{fig_intro}.
	Due to the existence of blow-up function $\varphi_1(\cdot)$ in $\varphi(\cdot,\cdot)$ and $\bm{b}_t$, showing the boundness of the guidance term $\bm{u}_t$ for all $t \in [0,1)$ is also essential, such that the similarity to the target distribution and numerical stability are ensured.
	The boundness of $\bm{u}_t$ is shown in the following proposition.
	
	\begin{proposition} \label{proposition_bounded_u}
		Suppose $\mathrm{d} h(\bm{s}) / \mathrm{d} \bm{s} \ne \bm{0}_{d_s \times 1}$, $\forall \bm{s} \in \mathbb{S}$ and \cref{assumption_safe_set} is satisfies, and use the blow-up function $\varphi_1(\cdot)$ in exponential form or inverse polynomial form as \eqref{eqn_varphi_1_exponential} and \eqref{eqn_varphi_1_polynomial} with $\omega > 2$, respectively.
		If the value of $\bm{v}^{\bm{\theta}}_t(\bm{\mathcal{T}}_0)$ is bounded, then the guidance term $\bm{u}_t$ is also bounded for all $t \in [0,1)$.
	\end{proposition}
	 \begin{IEEEproof}
	 	According to the closed-form of $\bm{u}_t^k$ in \eqref{eqn_u_closed_form}, the boundness of $\bm{u}_t$ depends on finite $a_t^k$, which is also related to $\| \bm{v}^{\bm{\theta}}_t(\bm{\mathcal{T}}_0) \|$ and $| \varphi(t, h(\bm{s}_t^k)) h(\bm{s}_t^k) |$.
	 	While the boundness of $\bm{v}^{\bm{\theta}}_t(\bm{\mathcal{T}}_0)$ is assumed, the upper bound of $| \varphi(t, h(\bm{s}_t^k)) h(\bm{s}_t^k) |$ is desired to derive.
	 	First, we show the derivative of the blow-up function $\varphi_1(\cdot)$ as
	 	\begin{align}
	 		\dot{\varphi}_1(t) = \exp(- \omega (1 - t)) \varphi_1^2(t)
	 	\end{align}
	 	for exponential form \eqref{eqn_varphi_1_exponential} and
	 	\begin{align}
	 		\dot{\varphi}_1(t) = 2 \omega^{-1} \varphi_1^2(t)
	 	\end{align}
	 	with $\omega \in \mathbb{R}_+$ for inverse polynomial form, which is then summarized as $\dot{\varphi}_1(t) \le c_{\varphi}(t) \varphi_1^2(t)$ with constant $c_{\varphi}(t) = \max\{\exp(- \omega (1 - t)), 2 \omega^{-1}\} \in \mathbb{R}_+$.
	 	Due to the choice of $\omega > 2$, it is easy to see $c_{\varphi}(t) < 1$ for all $t \in [0,1)$.
	 	Considering the blow-up function is only used when $h(\bm{s}_t^k) < 0$, the boundness of $\varphi(t, h(\bm{s}_t^k)) h(\bm{s}_t^k)$ is equivalent to the bounded $\varphi_1(t) h(\bm{s}_t^k)$ with $h(\bm{s}_t^k) < 0$, which is proven using Lyapunov theory.
	 	Specifically, the Lyapunov candidate is chosen as
	 	\begin{align}
	 		V_t^k = \varphi_1^2(t) h^2(\bm{E}_k \bm{\mathcal{T}}_t) / 2,
	 	\end{align}
	 	whose derivative is written as
	 	\begin{align}
	 		\dot{V}_t^k =& \varphi_1(t) \dot{\varphi}_1(t) h^2(\bm{E}_k \bm{\mathcal{T}}_t) - \varphi_1^2(t) | h(\bm{E}_k \bm{\mathcal{T}}_t) | \dot{h}(\bm{E}_k \bm{\mathcal{T}}_t) \nonumber \\
	 		\le& c_{\varphi}(t) \varphi_1^3(t) h^2(\bm{E}_k \bm{\mathcal{T}}_t) - \varphi_1^3(t) h^2(\bm{E}_k \bm{\mathcal{T}}_t) \\
	 		=& - 2 (1 - c_{\varphi}(t)) \varphi_1(t) V_t^k \nonumber
	 	\end{align}
	 	considering $h(\bm{E}_k \bm{\mathcal{T}}_t) < 0$.
	 	Following the variation of constants formula and the comparison principle, the value of $\dot{V}_t^k$ is lower bounded by
	 	\begin{align}
	 		\dot{V}_t^k \le \exp\Big(- 2 \int_0^t (1 - c_{\varphi}(s)) \varphi_1(s) \mathrm{d} s \Big) V_0^k \le V_0^k,
	 	\end{align}
	 	which indicates the bounded $| \varphi_1(t) h(\bm{E}_k \bm{\mathcal{T}}_t) |$ by
	 	\begin{align}
	 		\varphi_1(t) | h(\bm{E}_k \bm{\mathcal{T}}_t) | \le \varphi_1(0) | h(\bm{E}_k \bm{\mathcal{T}}_0) |
	 	\end{align}
	 	and therefore the bounded $a_t^k$, concluding the proof.
	 \end{IEEEproof}

	While the proposed FMBF-based safe flow matching \eqref{eq_controlled_dynamics} ensures the safety of all generated trajectories, the practical application relies heavily on $h(\cdot)$.
	However, in practice, finding such a function $h(\cdot)$ describing the safe set $\mathbb{C}_s$ is difficult, especially in complicated scenarios with multiple obstacles.
	In the next subsection, an improvement is made to handle the complicated safe set $\mathbb{C}_s$.
	
	\subsection{Safe Flow Matching under Composite Constraints}
	
	For a complex safe set, defining a safe set using a single FMBF may be inadequate for capturing intricate or multiple safety constraints. 
	In this subsection, we introduce a composite safe set defined by $N \in \mathbb{N}_{> 0}$ differentiable continuous functions $h_j(\cdot): \mathbb{S} \to \mathbb{R}$ with $j = 1, \cdots, N$.
	Then, a composite safe set $\mathbb{C}_s$ is defined as
	\begin{align} \label{eqn_composite_safe_set}
		\mathbb{C}_s = \bigcap\nolimits_{j = 1}^N \mathbb{C}_{s,j}, && \text{with} &&\mathbb{C}_{s,j} = \{ \bm{s} \in \mathbb{S} : h_j(\bm{s}) \ge 0 \}
	\end{align}
	and $j = 1, \cdots, N$.
	Moreover, we extend the definition of safe trajectory, safe flow matching in \cref{definition_safe_flow_matching}, and FMBF in \cref{definition_flow_matching_barrier_function} from the unified case to this composite framework, as detailed below.
	
	\begin{definition} [Composite Safe Trajectory and Composite Safe Flow Matching] \label{definition_safe_flow_matching_complex}
		A trajectory $\bm{\mathcal{T}}$ of length $H+1$ is safe for composite constraints $\mathbb{C}_s$ in \eqref{eqn_composite_safe_set}, if it satisfies the condition in \cref{definition_safe_flow_matching} w.r.t barrier functions $h_j(\cdot)$ for all $j = 1, \cdots, N$.
		Moreover, a FM process is called composite safe w.r.t. $\bm{h}(\cdot)$, if all generated trajectories $\bm{\mathcal{T}} = \bm{\mathcal{T}}_1$ at $t = 1$ from \eqref{eq_controlled_dynamics} achieves composite safety for any initial $\bm{\mathcal{T}}_0 \sim p_0$.
	\end{definition}
	
	To address the safe trajectory generation problem under composite constraints, the proposed FMBF in \cref{definition_flow_matching_barrier_function} is extended as follows.
	
	\begin{definition} [Composite Flow Matching Barrier Function] \label{definition_flow_matching_barrier_function_complex}
		The vector function $\bm{h}(\cdot) = [h_j(\cdot)]_{j=1,\cdots,N}$ is called a composite flow matching barrier function (CFMBF), if
		\begin{align}
			\frac{\mathrm{d} h_j(\bm{E}_k \bm{\mathcal{T}}_t)}{\mathrm{d} \bm{E}_k \bm{\mathcal{T}}_t} \bm{E}_k \bm{v}_t^{\bm{\theta}}(\bm{\mathcal{T}}_0)  &+ \frac{\mathrm{d} h_j(\bm{E}_k \bm{\mathcal{T}}_t)}{\mathrm{d} \bm{E}_k \bm{\mathcal{T}}_t} \bm{u}_t^k \\
			&+ \varphi_j(t, h_j(\bm{E}_k \bm{\mathcal{T}}_t)) h_j(\bm{E}_k \bm{\mathcal{T}}_t) \ge 0 \nonumber
		\end{align}
		for all $k = 0, \cdots, H$ and $t \in [0,1)$.
		The functions $\varphi_j(\cdot, \cdot): [0,1) \times \mathbb{R} \to \mathbb{R}_{> 0}$ for all $j = 1, \cdots, N$ are selected as
		\begin{align}
			\varphi_j(t,h) = \begin{cases}
				\varphi_{0,j}, & \text{if}~ h \ge 0 \\
				\varphi_{1,j}(t), & \text{otherwise}
			\end{cases},
		\end{align}
		where $\varphi_{0,j} \in \mathbb{R}_{> 0}$ and $\varphi_{1,j}(\cdot): [0,1) \to \mathbb{R}_{> 0}$ are chosen as monotonically increasing and satisfying $\int_0^{1^-} \varphi_{1,j}(s) \mathrm{d} s = \infty$.
	\end{definition}
	
	The function $\varphi_j(\cdot,\cdot)$ can be chosen e.g., in exponential form or inverse polynomial form as shown in \eqref{eqn_varphi_1_exponential} and \eqref{eqn_varphi_1_polynomial}, respectively.
	Moreover, building on the safety guarantees established for a unified safe set, we demonstrate that a CFMBF $\bm{h}(\cdot)$ ensures the safety of both the trajectory and the FM process in the composite case, as formalized below.
	
	\begin{theorem}
		\label{theorem_FMBF2}
		Suppose $\bm{h}(\cdot)$ is a valid CFMBF as defined in \cref{definition_flow_matching_barrier_function_complex}. 
		Choose the regulation term $\bm{u}_t$ from $\mathbb{U}_t = \prod_{k = 0}^H \mathbb{U}_t^k$ with $\mathbb{U}_t^k =\bigcap_{j = 1}^N \mathbb{U}_{j,t}^k$ and
		\begin{align}
			\mathbb{U}_{j,t}^k = \big\{ \bm{u}_t^k \in& \mathbb{R}^{d_s} \big| \frac{\mathrm{d} h_j(\bm{E}_k \bm{\mathcal{T}}_t)}{\mathrm{d} \bm{E}_k \bm{\mathcal{T}}_t} \bm{E}_k \bm{v}_t^{\bm{\theta}}(\bm{\mathcal{T}}_0) + \frac{\mathrm{d} h_j(\bm{E}_k \bm{\mathcal{T}}_t)}{\mathrm{d} \bm{E}_k \bm{\mathcal{T}}_t} \bm{u}_t^k \nonumber \\
			&+ \varphi_j(t, h_j(\bm{E}_k \bm{\mathcal{T}}_t)) h_j(\bm{E}_k \bm{\mathcal{T}}_t) \ge 0\big\}
		\end{align}
		for all $j = 1, \cdots, N$.
		Then, the flow matching process \eqref{eq_controlled_dynamics} is composite safe according to \cref{definition_safe_flow_matching_complex} w.r.t $\bm{h}(\cdot)$, and all generated trajectories $\bm{\mathcal{T}}_1$ satisfy the composite safety conditions in \cref{definition_safe_flow_matching_complex}.
	\end{theorem}
	\begin{IEEEproof}
		Utilizing the result in \cref{theorem_FMBF}, it is straightforward to see the existence of $\bm{u}_t^k$ for any $k = 0, \cdots, H$ and $t \in [0,1)$, such that trajectory $\bm{\mathcal{T}}_1$ is safe w.r.t $h_j(\cdot)$ as in \cref{definition_safe_flow_matching_complex} for all $j = 1, \cdots, N$ irrespective of the initial trajectory $\bm{\mathcal{T}}_0$.
		Consequently, the flow matching process meets the requirements for composite safe flow matching as defined in \cref{definition_safe_flow_matching_complex}, completing the proof.
	\end{IEEEproof}
	
	Same as the scenario in the unified contracts, we incorporate a minimal regularization term $\bm{u}_t$ into the flow matching process while ensuring safety under composite constraints. 
	For SafeFlow under composite constraints, the QP problem is formulated to determine $\bm{u}_t$ for any $t \in [0,1)$ as
	\begin{subequations}
		\label{eq_qp_cc}
		\begin{align}
			&\bm{u}_t = \argmin\nolimits_{\bm{u} = [(\bm{u}^0)^{\top}, \cdots, (\bm{u}^H)^{\top}]^{\top} \in \mathbb{R}^d} \| \bm{u} \|^2 \\
			\text{s.t.~}& a_t^{j,k}  + (\bm{b}_t^{j,k})^{\top} \bm{u}^k \ge 0, \forall k = 0, \cdots, H, \forall j = 1, \cdots, N. \nonumber
		\end{align}
	\end{subequations}
	where $a_t^{j,k} \in \mathbb{R}$ and $\bm{b}_t^{j,k} \in \mathbb{R}^{d_s}$ are written as
	\begin{align}
		&\bm{b}_t^{j,k} \!\!=\!\! (\partial h_j(\bm{E}_k \bm{\mathcal{T}}_t) / \partial (\bm{E}_k \bm{\mathcal{T}}_t))^{\top}, \\
		&a_t^{j,k} \!\!=\!\! (\bm{b}_t^{j,k})^{\top} \bm{E}_k \bm{v}_t^{\bm{\theta}}(\bm{\mathcal{T}}_0) \!\!+\!\! \varphi_j(t, h_j(\bm{E}_k \bm{\mathcal{T}}_t)) h_j(\bm{E}_k \bm{\mathcal{T}}_t).
	\end{align}	
	The problem in \cref{eq_qp_cc} can be decoupled into independent subproblems for each $\bm{u}_t^i$, such that
	\begin{align}
		&\bm{u}_t^k = \argmin\nolimits_{\bm{u}^k \in \mathbb{R}^{d_s}} \| \bm{u}^k \|^2 \nonumber\\
		\text{s.t.~}& a_t^{j,k}  + (\bm{b}_t^{j,k})^{\top} \bm{u}^k \ge 0, \forall j \in \{1, \cdots, N\}. \nonumber
	\end{align}
	However, with multiple linear constraints, the feasibility of this decoupled QP is not guaranteed.
	To ensure feasibility, we introduce relaxation terms $\delta_j^k \in \mathbb{R}_{\ge 0}$ and reformulate decoupled QP problem as
	\begin{align}
		&\bm{u}_t^k = \argmin\nolimits_{\bm{u}^k \in \mathbb{R}^{d_s}, \delta_j^k \in \mathbb{R}_{\ge 0}} \| \bm{u}^k \|^2 + \sum\nolimits_{j = 1}^N {\delta_j^k}^2 \nonumber \\
		\text{s.t.~}& a_t^{j,k}  + (\bm{b}_t^{j,k})^{\top} \bm{u}^k + \delta_j^k \ge 0, \forall j \in \{1, \cdots, N\}. \nonumber
	\end{align}
	It is straightforward to verify that choosing sufficiently large values for $\delta_j^i$ always ensures the feasibility of the QP problem. 
	This effectively turns the problem into a feasibility problem.
	For instance, one feasible solution is $\bm{u}_i = \bm{0}_{d_s \times 1}$ and $\delta_j^i = \max\{0, - \frac{\partial h_j(\bm{E}_k \bm{\mathcal{T}}_t)}{\partial \bm{E}_k \bm{\mathcal{T}}_t} \bm{E}_k \bm{v}_t^{\bm{\theta}}(\bm{\mathcal{T}}_0) - \varphi_j(t, h_j(\bm{E}_k \bm{\mathcal{T}}_t)) h_j(\bm{E}_k \bm{\mathcal{T}}_t) \}$ for all $j = 1, \cdots, N$.
	There also exist other feasibility guaranteed methods for multiple constraints in CBF-based methods, which are not discussed in detail here due to the main scope of this paper lying in the safe trajectory generation framework.
	
	\begin{remark}	
		Due to such relaxation for feasibility, the positivity for $h_j(\bm{E}_k \bm{\mathcal{T}}_{1^-})$ may not be satisfied.
		In such cases with additional consideration of the numerical stability, the deployment of the terminal safety filter as shown in \cref{fig_intro} is essential.
		Specifically, the safe trajectory $\bm{\mathcal{T}} = \bm{\mathcal{T}}_1$ is obtained from the terminal filter by solving
		\begin{align} \label{eqn_T1_optimization_complex}
			&\bm{\mathcal{T}}_1 = \argmin\nolimits_{\bm{\mathcal{T}} \in \mathbb{R}^d} \| \bm{\mathcal{T}} - \bm{\mathcal{T}}_{1^-} \| \\
			\text{s.t.~}& h_j(\bm{E}_k \bm{\mathcal{T}}) \geq 0, ~\forall k = 0, \cdots, H, \forall j = 1, \cdots, N, \nonumber
		\end{align}
		whose feasibility is guaranteed if $\mathbb{C}_s$ is non-empty.
		The non-empty $\mathbb{C}_s$ indicates the existence of $\bm{s}_{safe} \in \mathbb{C}_s \subseteq \mathbb{S}$.
		Then, the trajectory $\bm{\mathcal{T}}_{safe} = [\bm{s}_{safe}^{\top}, \cdots, \bm{s}_{safe}]^{\top}$ is safe as in \cref{definition_safe_flow_matching_complex}, which indicates the existence of at least one solution, i.e., $\bm{\mathcal{T}}_1 = \bm{\mathcal{T}}_{safe}$, for \eqref{eqn_T1_optimization_complex}.
	\end{remark}
	
	\section{Experimental Results}
	\label{sec_result}	
	
	In this section, we conduct experiments\footnote{More details are shown in Appendix B.} to answer the following questions:
	\begin{itemize}
		\item [\textbf{Q1}:] Does SafeFlow generate obstacle-avoiding trajectories with provable safety guarantees? 
		\item [\textbf{Q2}:] What impact do FMBFs have on the diffusion and flow-based model in the inference process? 
		\item [\textbf{Q3}:] How does SafeFlow perform in comparison to SOTA methods that incorporate safety constraints? 
	\end{itemize}
	
	\subsection{General Experiment Settings} \label{subsection_general_simulation_setting}
	
	In this paper, two evaluation settings are considered, namely robot navigation in \cref{subsec_planar} and robot manipulation in \cref{subsec_experiment}, in which our proposed methods are compared with current existing generative model-based motion planning methods.
	In particular, we compare against several filter-based SOTA methods: 
	\begin{itemize}
		\item \textbf{SafeDiffuser}~\cite{xiao2025ICLRsafediffuser}, including all variants of  Robust-Safe Diffuser (RoSD), Relaxed-Safe Diffuser (ReSD) and Time-Varying-Safe Diffuser (TVSD);
		\item \textbf{CoB and CoBL}~\cite{Mizuta2024IROScoBLDiffusion}, which combine control barrier functions with or without Lyapunov functions within the diffusion framework.
	\end{itemize}
	Besides, some conventional guidance-based approaches are also compared, namely \textbf{Classifier Guidance (CG)}~\cite{Dhariwal2021NeurIPSDiffusion} and \textbf{Truncate (TRUNC)}~\cite{brockman2016openaigym}.
	Moreover, we implement our proposed FMBFs on top of the diffusion model, named \textbf{SafeDiffus}, to ensure safety guarantees.
	Compared to the SafeDiffuser in \cite{xiao2025ICLRsafediffuser}, the proposed SafeDiffus employs prescribed-time control barrier functions as in \cref{theorem_FMBF,theorem_FMBF2}, such that the safety is theoretically guaranteed with a probability of $1$.
	
	The experiments evaluate the performance of SafeFlow, SafeDiffus and other methods for planar robot navigation and manipulation with $5$ metrics, which are explained as follows. 
	\begin{itemize}[left=4pt,itemsep=0pt]
		\item \textbf{Safety Rate:} 
		Proportion of generated trajectories that avoid collisions with obstacles.
		\item \textbf{Inference Time:} 
		Average time required to generate a trajectory during test-time inference, reflecting the computational efficiency of the planner. 
		The time interval starts from the initial sampling until the trajectory is generated.
		\item \textbf{Kullback–Leibler Divergence:} 
		To quantify the discrepancy between the distribution of generated final states and the true distribution observed in the dataset, we employ the Kullback–Leibler (KL) divergence. 
		In particular, we characterize the impact of distribution discrepancy on target achievement by evaluating the final state $\bm{s}^H = \bm{E}_H \bm{\mathcal{T}}$ of the trajectory $\bm{\mathcal{T}}$.
		Collect all final states for $N \in \mathbb{N}$ generated trajectories in a discrete set $\{ (\bm{s}^H)_i \}_{i=1}^N$, and approximate the empirical distributions over generated final states as $Q$ and over dataset $\{ \bm{E}_H \bm{\mathcal{T}}^{(\iota)} \}_{\iota \in \mathbb{N}}$ as $P$ using Gaussian kernel density estimation \cite{heer2021fast}.
		With the approximated distributions, the KL divergence between $P$ and $Q$ is computed as
		\begin{align}
			D_{\mathrm{KL}}(P \| Q) = \frac{1}{N} \sum\nolimits_{i=1}^N P((\bm{s}^H)_i) \log \frac{Q((\bm{s}^H)_i)}{P((\bm{s}^H)_i)}.
		\end{align} 
		Note that a lower KL divergence indicates closer alignment between the generated state distribution and the target distribution in the dataset.		
		\item \textbf{Curvature-based Smoothness (CS)}: 
		To quantify the smoothness of a trajectory, we employ a curvature-based smoothness metric that captures the degree of directional change between consecutive path segments. 
		Given a sequence of trajectory points $\{\bm{s}^k\}_{k=0}^H$, the segment vectors are defined as $\bm{w}^k = \bm{s}^k - \bm{s}^{k-1}$ for $k = 1, \cdots, H$. 
		The sine of the turning angle $\theta^k \in [0, \pi] \subset \mathbb{R}$ at each intermediate point is computed as 
		\begin{align}
			{\theta}^k = \cos^{-1} \big( (\bm{w}^k)^{\top} \bm{w}^{k+1} / (\| \bm{w}^k \| \| \bm{w}^{k+1} \|) \big)
		\end{align}
		with $k = 1, \cdots, H - 1$, where the numerator denotes the magnitude of the cross product. The overall smoothness score $S_c \in \mathbb{R}_{\ge 0}$ is then defined as the average of these sine values over the trajectory as
		\begin{align}
			S_c = (H - 1)^{-1} \sum\nolimits_{k=1}^{H-1} (1 - \cos \theta^k).
		\end{align}
		It is noted that a lower absolute score $| S_c |$ indicates smoother transitions with minimal curvature, while higher values reflect abrupt changes in direction.
		\item \textbf{Acceleration-based Smoothness (AS)}: 
		To complement curvature-based measures, an acceleration-based smoothness metric is also introduced, which quantifies the change in velocity along the trajectory. 
		Given a sequence of state $\{\bm{s}^k\}_{k=0}^H$, the second-order differences are defined as 
		\begin{align}
			\bm{a}^k = \bm{w}^{k+1} - \bm{w}^k = \bm{s}^{k+1} - 2\bm{s}^k + \bm{s}^{k-1}
		\end{align}
		with $k = 1, \cdots, H-1$ to reflect the acceleration at step $k$. 
		The norm $\| \bm{a}^k \|$ captures the local acceleration, and the overall acceleration-based smoothness $S_a \in \mathbb{R}_{\ge 0}$ is defined as the average magnitude over the trajectory:
		\begin{align}
			S_a = (H - 2)^{-1} \sum\nolimits_{i=2}^{H-1} \|\bm{a}^i\|.
		\end{align}
		Smaller values of $S_a$ indicate a smoother trajectory with less abrupt changes in velocity, making it particularly relevant in applications where continuity and comfort of motion are critical.
	\end{itemize}
	
	To illustrate the inference procedure of our proposed SafeFlow approach in a practical way, we present the pseudocode in \cref{alg:otcfm_cbf} for the numerical integration. 
	To reduce the computational load introduced by the optimization in \eqref{eq_qp_cc}, the integral over $t \in [0, 1)$ is divided into multiple subintervals.
	Specifically, the numerical integrator \emph{torchdiffeq} is used with mode $dropi5$, which determines the number and length of subintervals automatically.
	Moreover, we design SafeFlow to prioritize distribution formation in the early stages with small $t$ values, while gradually emphasizing constraint satisfaction as $t$ increases. 
	To this end, the CFMBF component is only activated in SafeFlow for $t \ge t^*$ with $t^* = 0.5$.
	Considering the flow function $\bm{v}^{\bm{\theta}}(\cdot, \cdot)$ is trained in a normalized space and the barrier functions $h(\cdot)$ is constructed in state space, the processes of normalization and denormalization are used as in \cref{alg:otcfm_cbf} to convert between the normalized trajectory $\hat{\bm{\mathcal{T}}}$ for FM and the corresponding state space trajectory $\bm{\mathcal{T}}$.
	
	\subsection{Robot Navigation}
	\label{subsec_planar}
	
\begin{figure}[t]
		\centering
		\begin{subfigure}[b]{0.15\textwidth}
			\includegraphics[width=\linewidth]{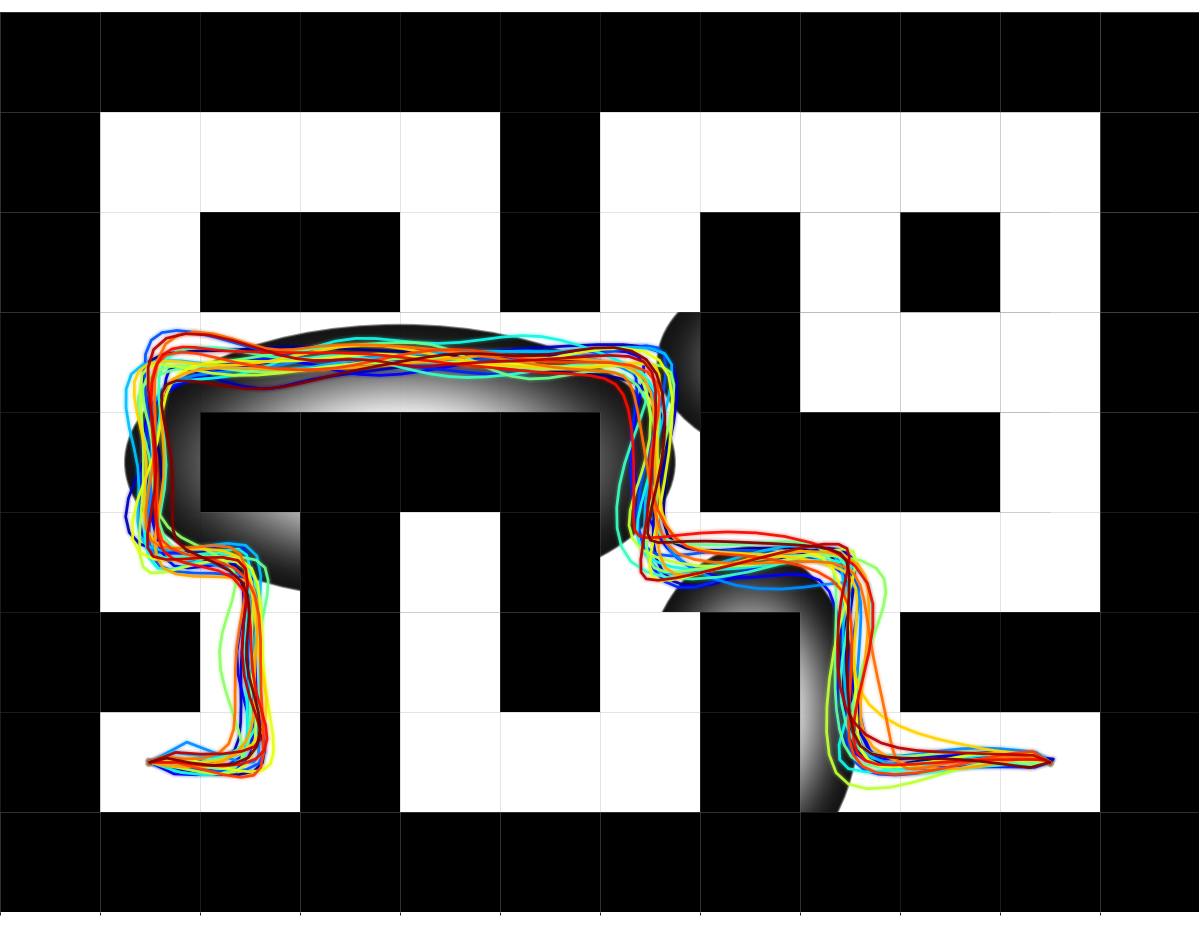}
			\caption{Dataset}
			\label{fig:training}
		\end{subfigure}
		\hfill
		\begin{subfigure}[b]{0.15\textwidth}
			\includegraphics[width=\linewidth]{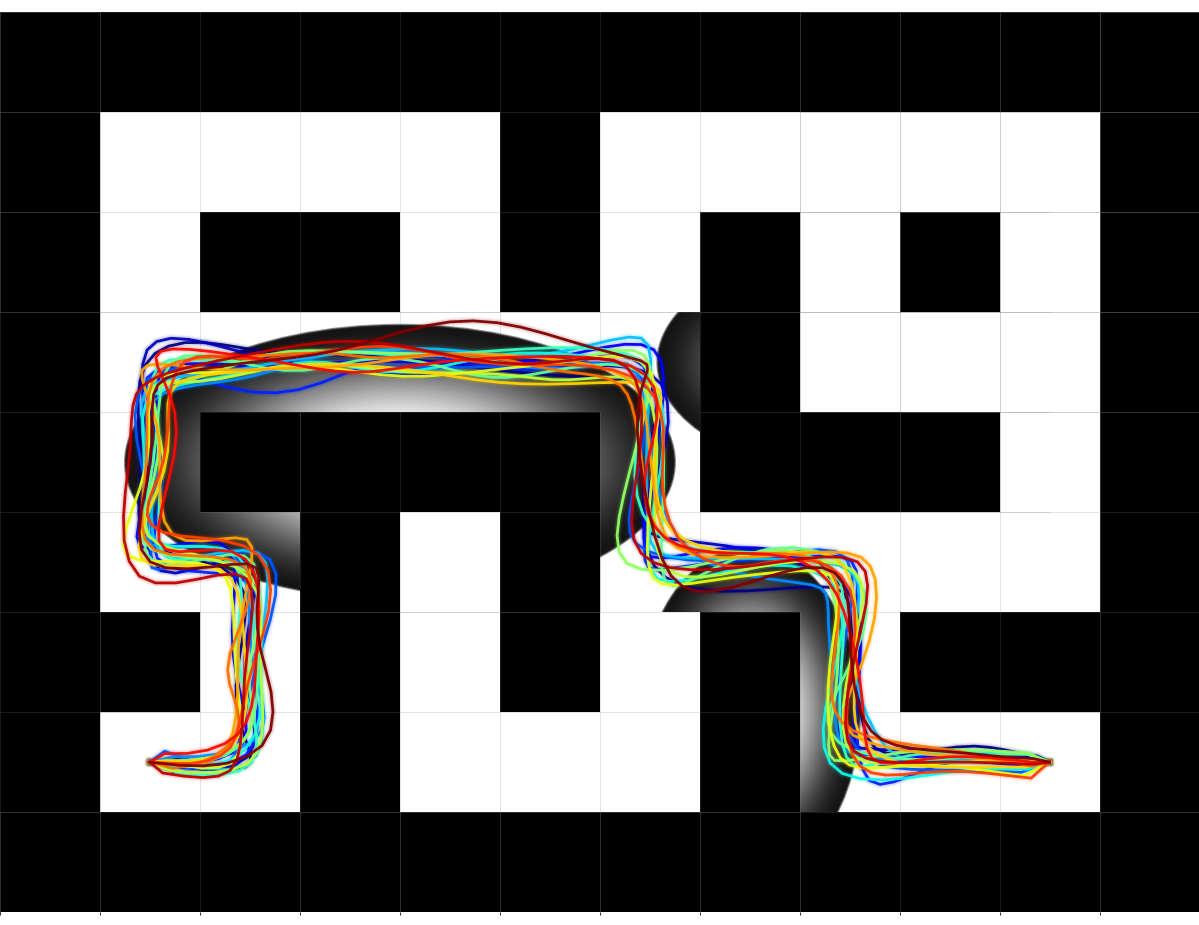}
			\caption{Diffuser}
		\end{subfigure}
		\hfill
		\begin{subfigure}[b]{0.15\textwidth}
			\includegraphics[width=\linewidth]{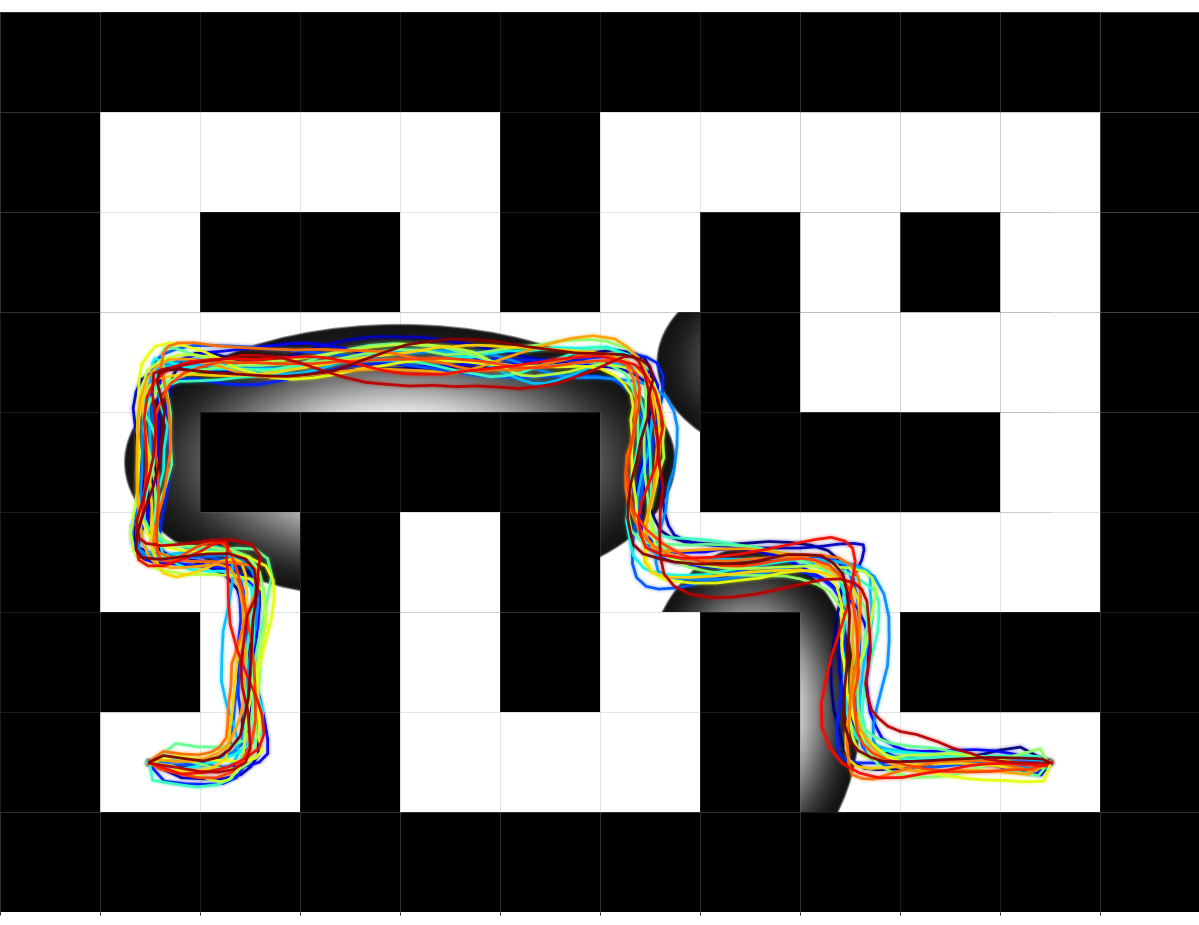}
			\caption{FM}
		\end{subfigure}
		
		\begin{subfigure}[b]{0.15\textwidth}
			\includegraphics[width=\linewidth]{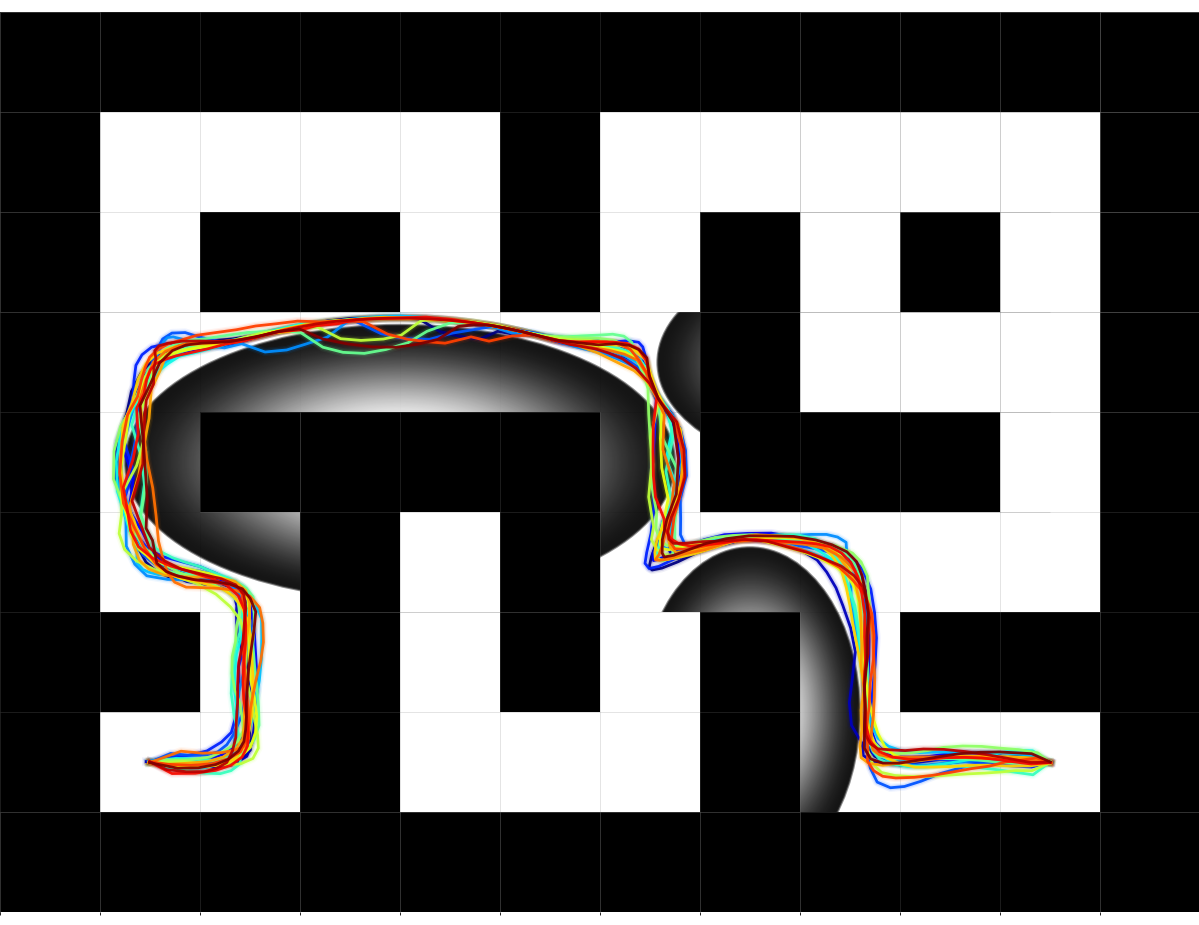}
			\caption{CG}
		\end{subfigure}
		\hfill
		\begin{subfigure}[b]{0.15\textwidth}
			\includegraphics[width=\linewidth]{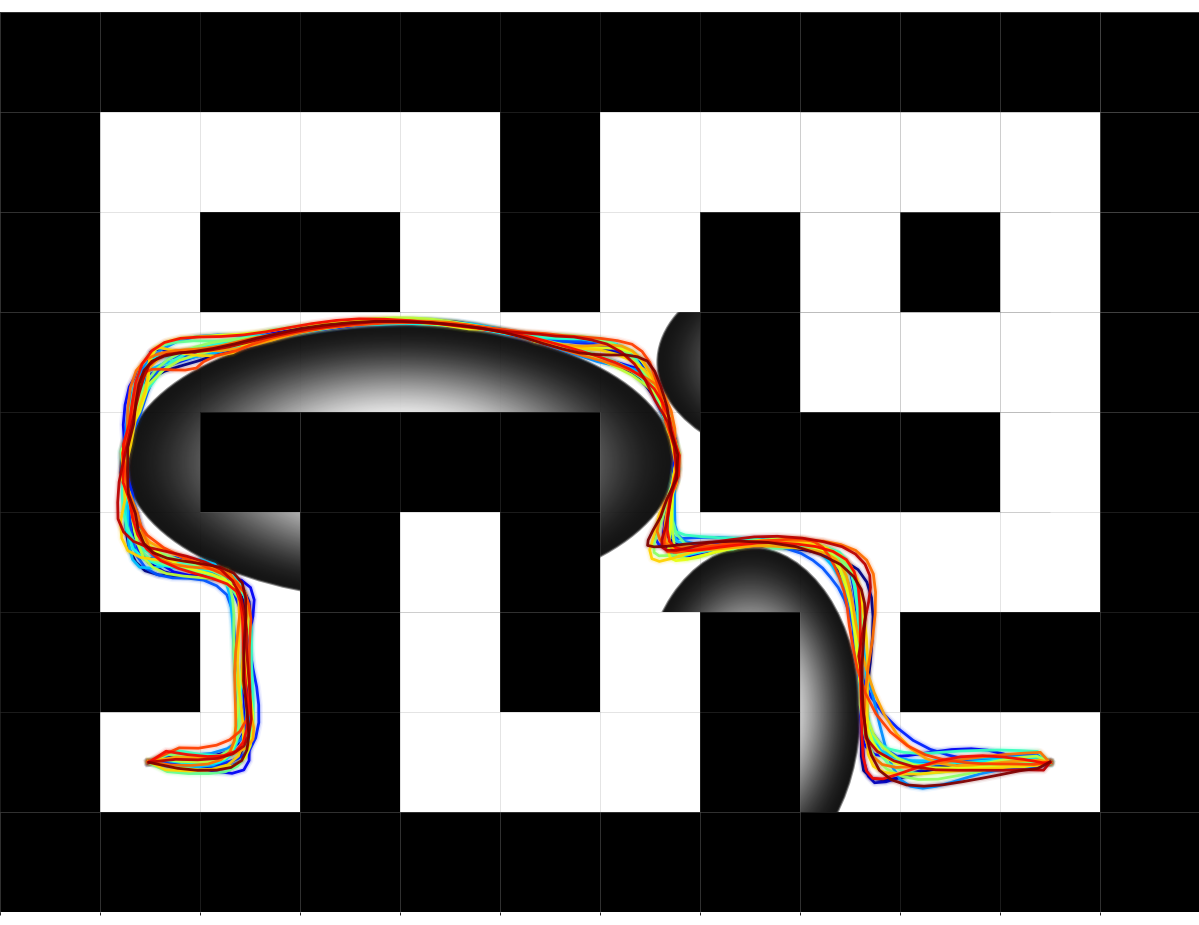}
			\caption{SafeDiffus (Ours)}
		\end{subfigure}
		\hfill
		\begin{subfigure}[b]{0.15\textwidth}
			\includegraphics[width=\linewidth]{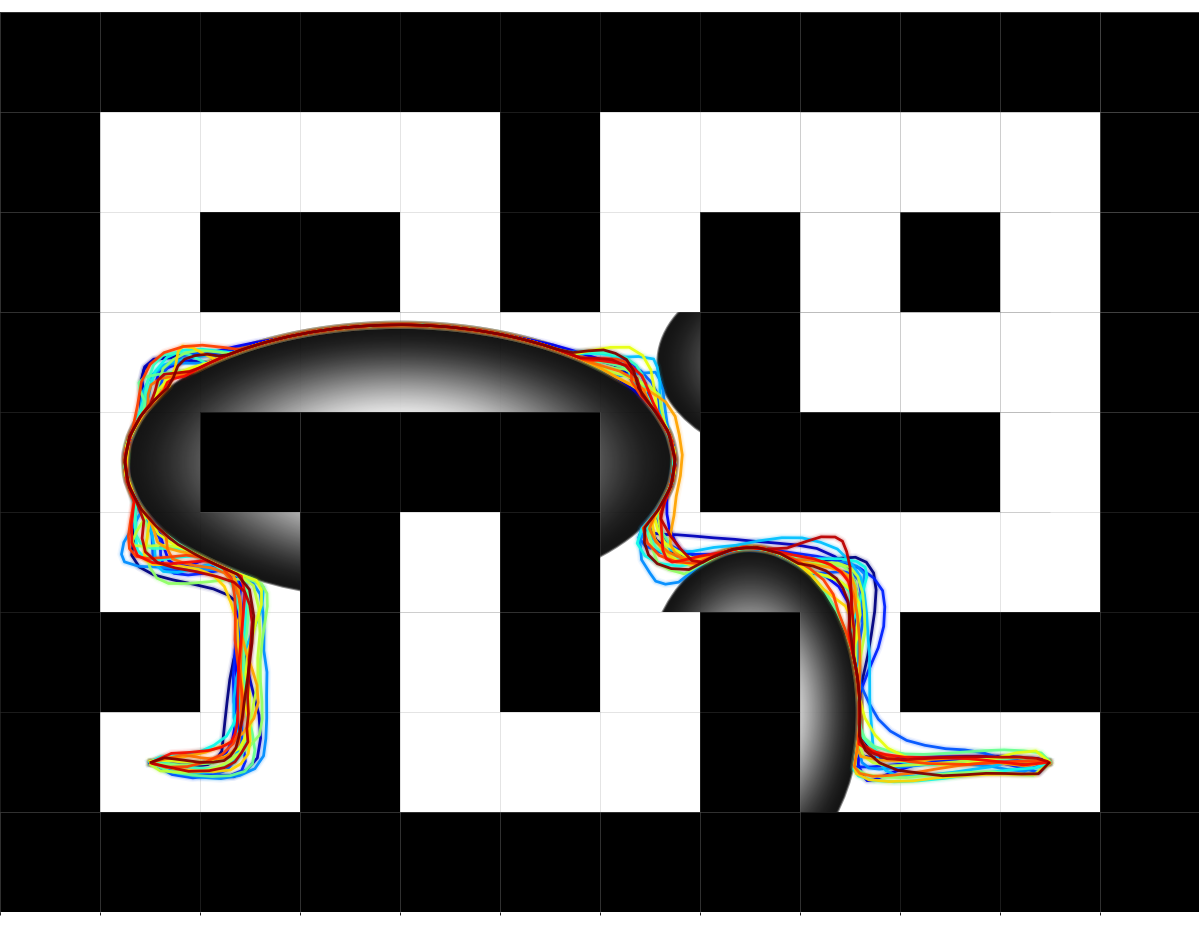}
			\caption{SafeFlow (Our)}
		\end{subfigure}
		
		\begin{subfigure}[b]{0.15\textwidth}
			\includegraphics[width=\linewidth]{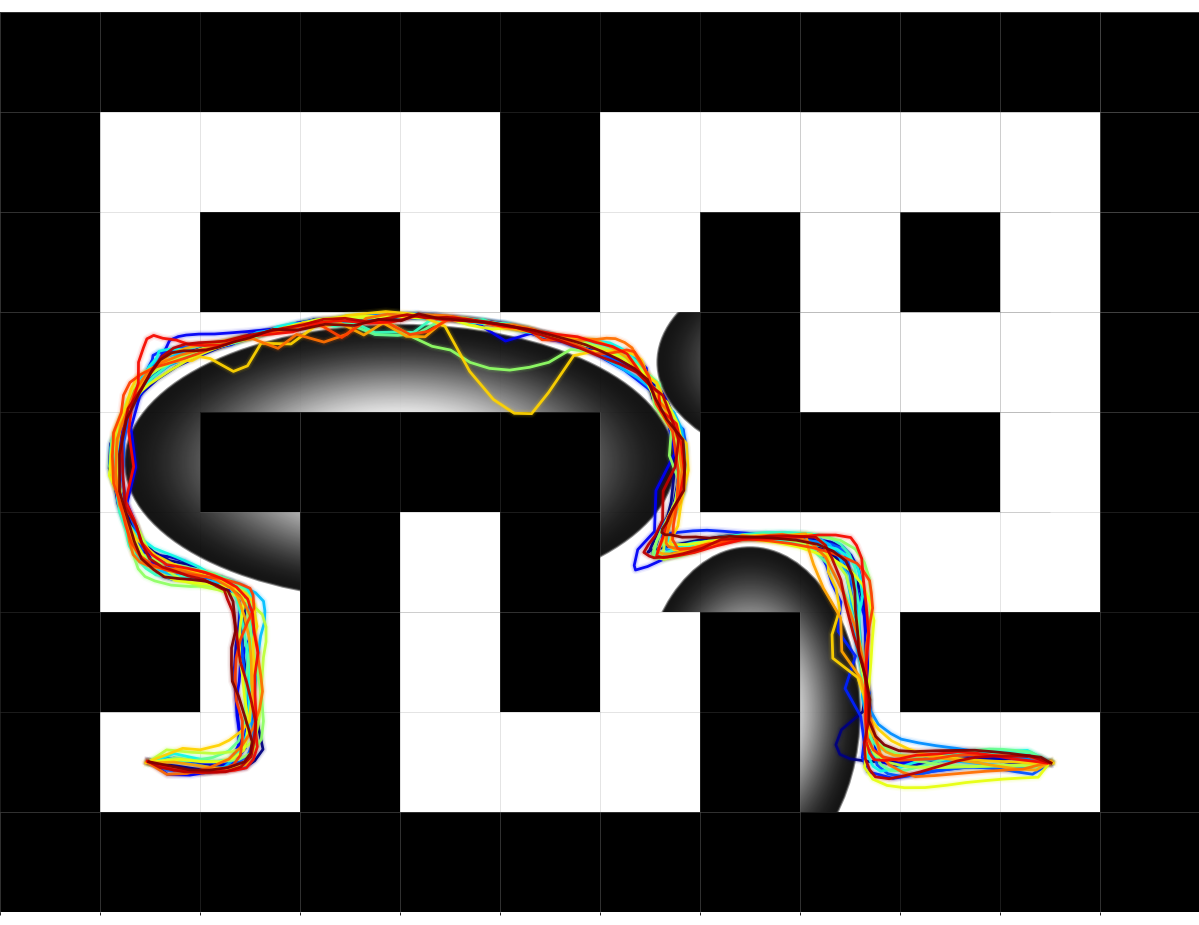}
			\caption{TRUNC}
		\end{subfigure}
		\hfill
		\begin{subfigure}[b]{0.15\textwidth}
			\includegraphics[width=\linewidth]{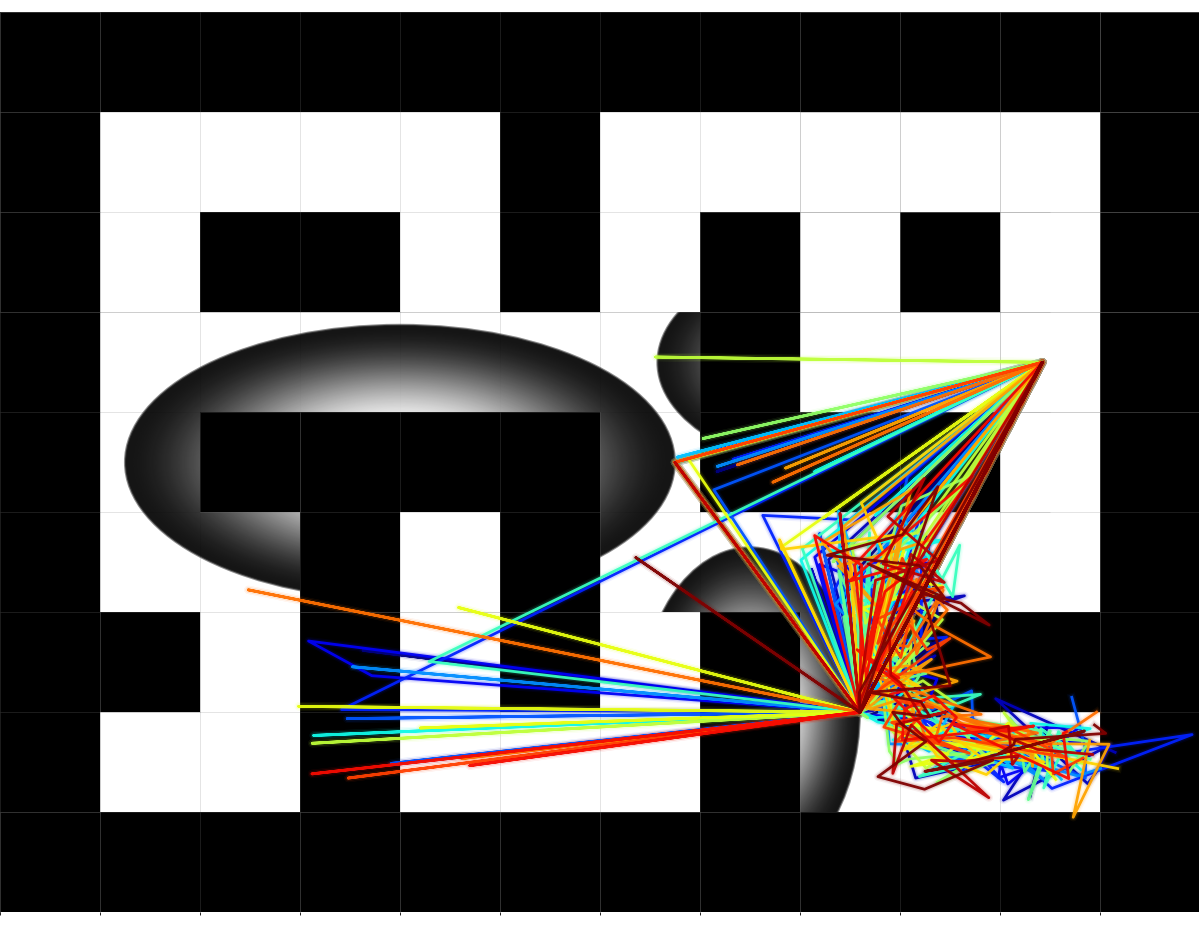}
			\caption{CoB}
		\end{subfigure}
		\hfill
		\begin{subfigure}[b]{0.15\textwidth}
			\includegraphics[width=\linewidth]{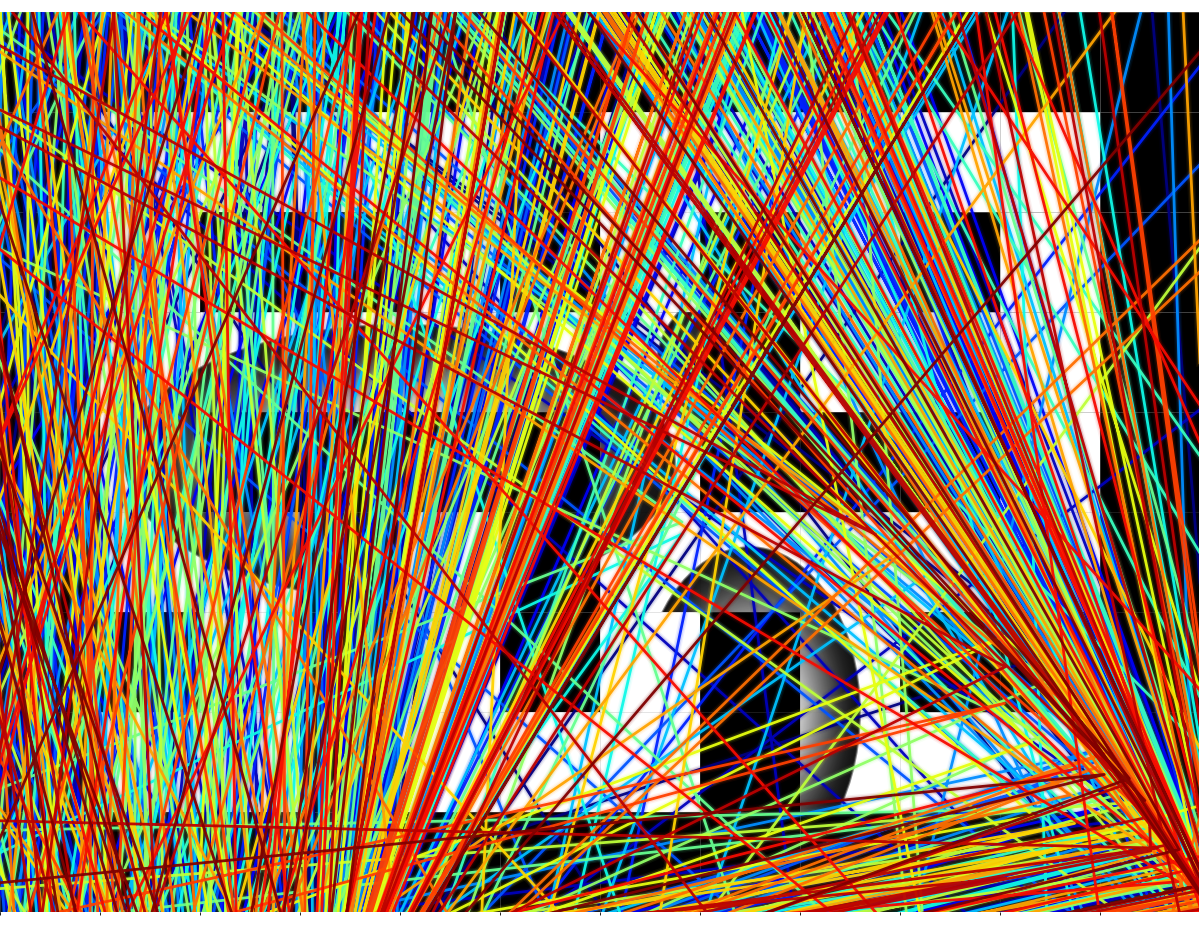}
			\caption{CoBL}
		\end{subfigure}
		
		\begin{subfigure}[b]{0.15\textwidth}
			\includegraphics[width=\linewidth]{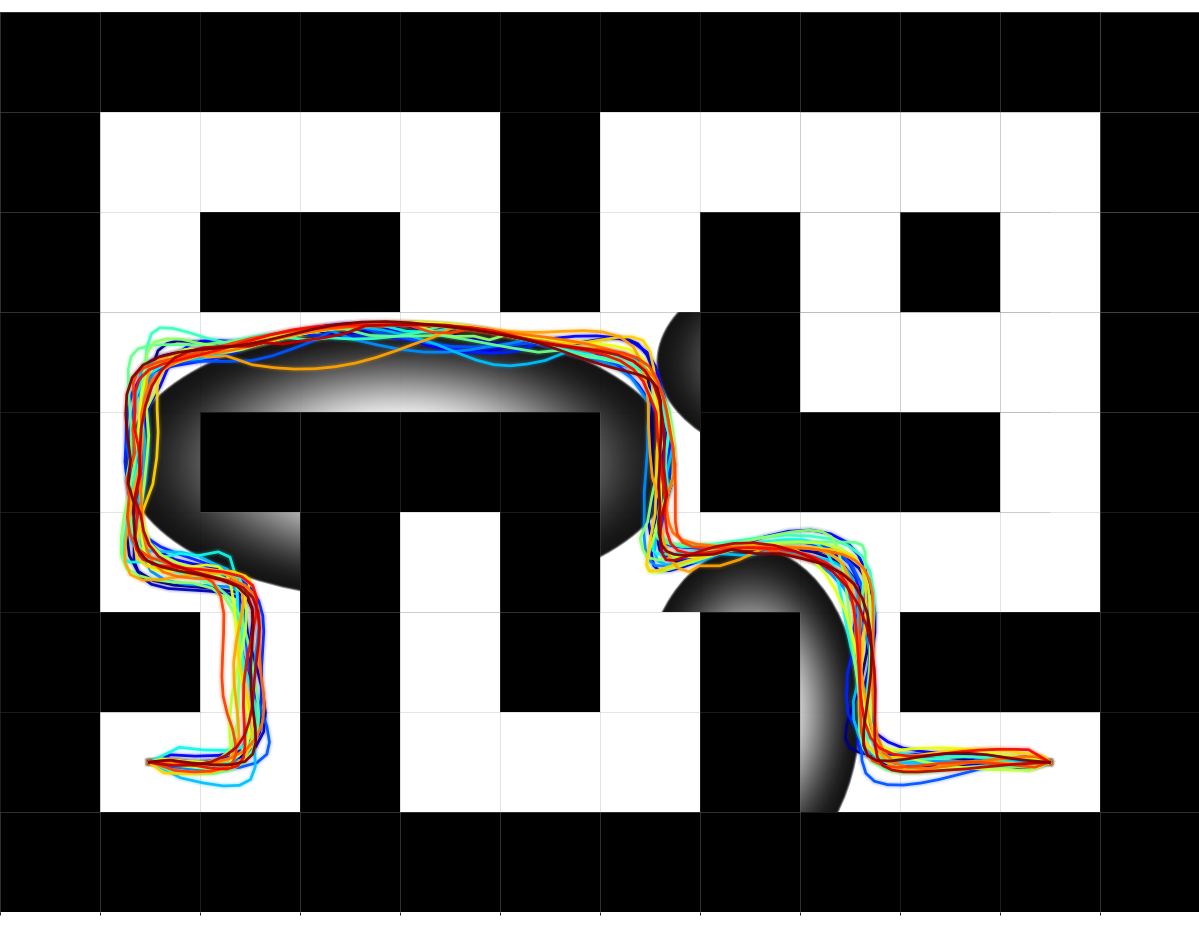}
			\caption{RoSD}
		\end{subfigure}
		\hfill
		\begin{subfigure}[b]{0.15\textwidth}
			\includegraphics[width=\linewidth]{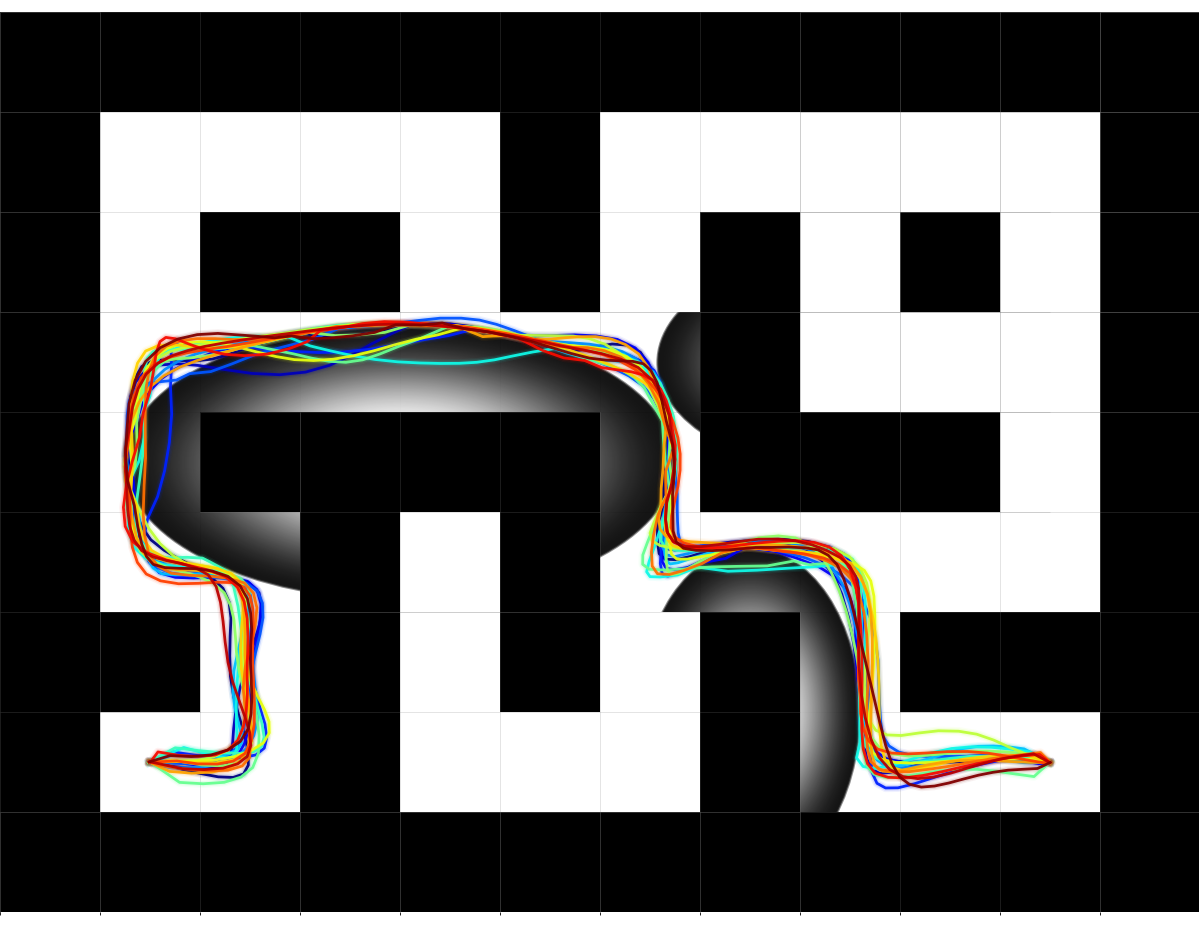}
			\caption{ReSD}
		\end{subfigure}
		\hfill
		\begin{subfigure}[b]{0.15\textwidth}
			\includegraphics[width=\linewidth]{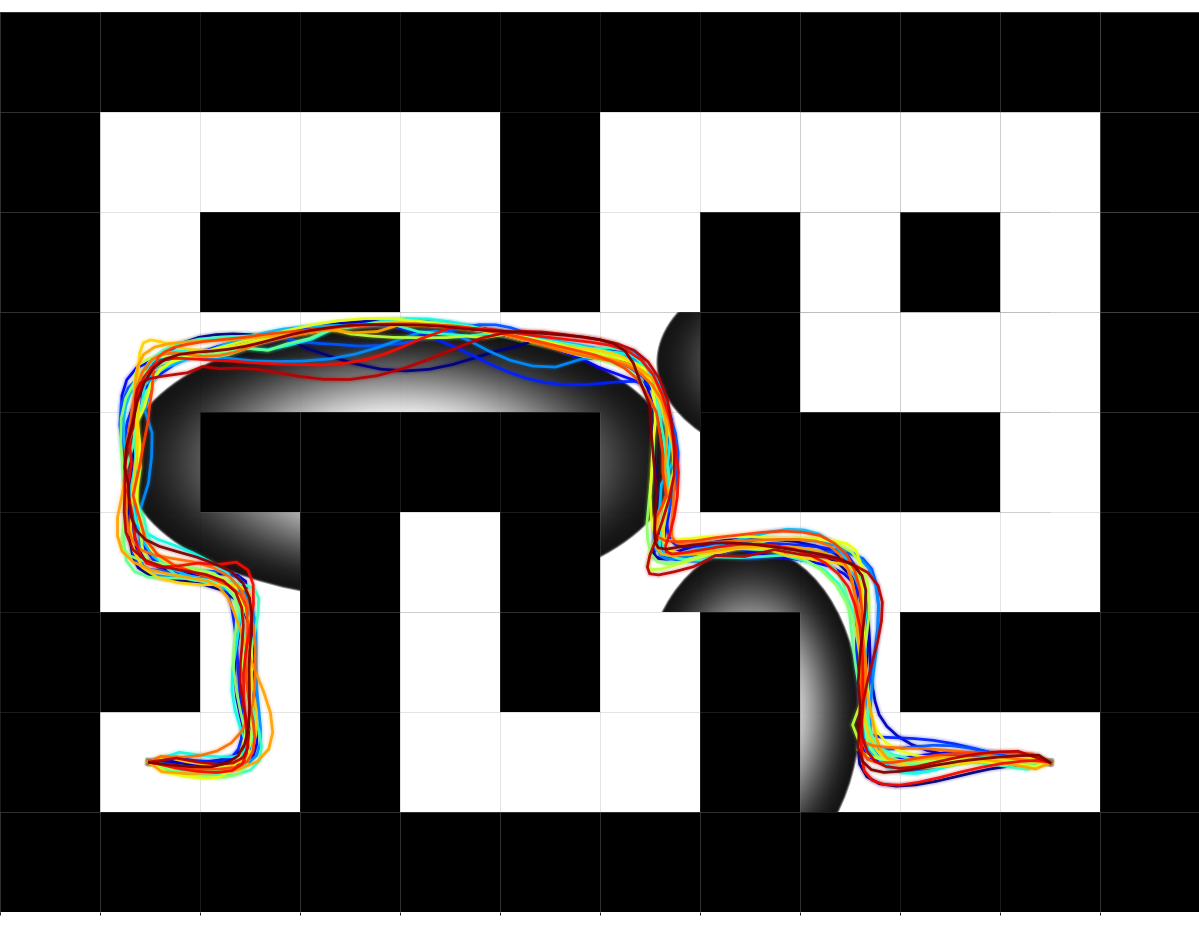}
			\caption{TVSD}
		\end{subfigure}
		\vspace{-0.3cm}
		\caption{Trajectories for robot navigation.}
		\vspace{-0.5cm}
		\label{fig_car12}
	\end{figure}

	In this subsection, we consider a simple task that the robot navigates in planar from the lower-left to lower-right in a maze, which  is illustrated in \cref{fig_car12}.
	The training data is generated following the same procedure in \cite{kim2022smooth}, which employs model predictive path integral (MPPI) to generate multiple smooth trajectories along a teaching trajectory.

	\begin{algorithm}[t]
		\caption{Inference Process of SafeFlow}
		\label{alg:otcfm_cbf}
		\begin{algorithmic}[1]
			\STATE Initialize normalizer and CFMBF scheduler
			\STATE $\hat{\bm{\mathcal{T}}}_{0}$ $\gets$ sample from $\mathcal{N}(\bm{0}_{d \times 1},\bm{I}_d)$
			\STATE $t \gets 0$; integration step $\varsigma \gets 0.001$;
			
			\WHILE{$t < 1$}
				\STATE $\hat{\bm{v}}^{\bm{\theta}}_t(\hat{\bm{\mathcal{T}}}_t) \gets$ learned flow with $t$ and $\hat{\bm{\mathcal{T}}}_t$;
				\STATE $\bm{v}^{\bm{\theta}}_t(\hat{\bm{\mathcal{T}}}_t)$ in state space $\gets$ denormalization of the normalized flow $\hat{\bm{v}}^{\bm{\theta}}_t(\hat{\bm{\mathcal{T}}}_t)$;
				\STATE $\bm{\mathcal{T}}_t$ in state space $\gets$ denormalization of the normalized trajectory $\hat{\bm{\mathcal{T}}}_t$;
				\STATE $\bm{u}_t \gets$ CFMBF-based QP \eqref{eq_qp_cc} with $t$, $\bm{\mathcal{T}}_t$ and $\bm{v}^{\bm{\theta}}_t(\hat{\bm{\mathcal{T}}}_t)$;
				\STATE $\hat{\bm{u}}_t \gets$ normalization of the input $\bm{u}_t$ in state space;
				
				\STATE $\bm{\mathcal{T}}_{\text{4th}}, \bm{\mathcal{T}}_{\text{5th}} \gets $ $4$-th, $5$-th order Runge-Kutta integration using learned $\bm{v}^{\bm{\theta}}(\cdot, \cdot) + \hat{\bm{u}}_t$ with $t$ and $\hat{\bm{\mathcal{T}}}_t$;
				\STATE Absolute error $\Delta_{\mathcal{T}}$, relative error $\Delta_{\mathcal{T},r} \!\gets\! \bm{\mathcal{T}}_{\text{4th}}, \bm{\mathcal{T}}_{\text{5th}}$;
				\IF{Both $\Delta_{\mathcal{T}}$ and $\Delta_{\mathcal{T},r}$ not exceed threshold}
					\STATE $t \gets t + \varsigma$; $\bm{\mathcal{T}} \gets \bm{\mathcal{T}}_{\text{5th}}$;
				\ENDIF
				\STATE Adjust $\varsigma$ based on $\varsigma$ and $\Delta_{\mathcal{T},r}$;

			\ENDWHILE
			
			\STATE $\bm{\mathcal{T}}_{1} \gets$ terminal CFMBF safety filter$(\bm{\mathcal{T}}_{1^-})$
		\end{algorithmic}
	\end{algorithm}

	Besides the constraints inherited from the maze itself, three additional obstacles are considered in ellipse form.
	Each ellipse obstacle $j=1, \cdots ,3$ is characterized with a center \(\bm{c}_j \in \mathbb{R}^2\) and semi-axes length \(a_j, b_j \in \mathbb{R}_+\), such that the safe set $\mathbb{C}$ is defined as complement of unsafe set $\bar{\mathbb{C}}$ with
	\begin{align}
		\bar{\mathbb{C}} = \bigcup\nolimits_{j = 1}^3 \mathbb{O}_j, && 
		\mathbb{O}_j = \{ \bm{s} \in \mathbb{R}^2 : h_j(\bm{x}) \le 0 \}
	\end{align}
	and $h_j(\bm{s}) := (\bm{s} - \bm{c}_j)^\top \bm{Q}_j (\bm{x} - \bm{c}_j) - 1$, $\bm{Q}_j = \mathrm{diag}(a_j^{-2}, b_j^{-2})$ for all obstacles with $j = 1, \cdots, 3$.
	The configuration of each obstacle is shown as
	\begin{align}
		\bm{c}_1 &= [3.5, 4.0]^{\top}, &
		\bm{c}_2 &= [8.0, 3.0]^{\top}, &
		\bm{c}_3 &= [7.0, 6.5]^{\top}, \\
		a_1 &= 2.5, & a_2 &= 1.75, & a_3 &= 1, \\
		b_1 &= 1.25, & b_2 &= 1, & b_3 &= 1.5.
	\end{align}
	Furthermore, the blow-up functions are defined as
	\begin{align}
		\varphi_0(t) = 1, && \varphi_1(t) = 
		\begin{cases}
			1 + 4 t^3, & \displaystyle {t} < \gamma, \\
			1 / (T - t), & \displaystyle {t} \ge \gamma,
		\end{cases}
	\end{align}
	with a threshold factor \(\gamma=0.9\). 
	
	To illustrate the effectiveness of SafeFlow in generating safe, smooth, and goal-directed trajectories for planar robot navigation, we qualitatively analyze the trajectory generation process based on benchmark experiments with $1000$ trials in \cref{tab_car}. 
	Notably, SafeFlow achieves a perfect safety rate of 100\%, significantly outperforming all baselines with safety guarantees, which provides both experimental and theoretical support for \textbf{Q1}. 
	It is noted that the safety claimed in \cite{Mizuta2024IROScoBLDiffusion,xiao2025ICLRsafediffuser} is only probabilistic due to the stochastic property of the diffusion model and the utilization of conventional CBF, which induces low safety in our scenarios.
	To answer \textbf{Q2}, we further analyze the differences between the vanilla FM and SafeFlow, as well as the vanilla Diffuser and SafeDiffus. 
	SafeFlow and SafeDiffus incur approximately 60\% higher inference time due to the safety regularization. 
	Nevertheless, they outperform other baselines that incorporate safety strategies. 
	The increases in KL and CS are expected, as the models need to make directional adjustments to avoid obstacles. 
	Notably, the lower CS values of FM and the vanilla Diffuser reflect overly simplistic, straight-line trajectories that compromise safety. 
	Interestingly, in the AS metric, our SafeDiffus outperforms the vanilla Diffuser, while FM and SafeFlow and FM yield the lowest AS scores. 
	The answer to \textbf{Q3} is illustrated in \cref{fig_car12}, which visually compares the trajectory generation performance across different methods.
	It is evident that SafeFlow effectively avoids obstacles while closely adhering to the trajectory distribution observed in the training dataset. 
	SafeDiffus also achieves zero collisions; however, some outliers remain, where a few generated trajectories exhibit higher sensitivity.
	In contrast, other approaches either collide with obstacles or generate trajectories that deviate significantly from the training distribution.
	This highlights the ability of the proposed framework to outperform SOTA methods, offering both training-free deployment and test-time safety guarantees. 
	
	\begin{table}[t]
		\centering
		\caption{Safe planning for robot navigation comparisons.}
		\label{tab_car}
		\begin{tabular}{l c c c c c}
			\hline 
			\textbf{Method} & $\!\!$\textbf{Safety$\uparrow\!$ }  &$\!\!$\textbf{KL$\downarrow\!$} & $\!\!$\textbf{CS$\downarrow\!$ } & $\!\!$\textbf{AS$\downarrow\!$} & $\!\!$\textbf{Time$(\!s\!)\!\downarrow\!$} \\
			\hline
			
			CG~\cite{Dhariwal2021NeurIPSDiffusion}    & 10.39\%  & 0.0252 & 0.2492 & 0.0152 & 0.96\\
			
			TRUNC~\cite{brockman2016openaigym}    & 11.24\%  & 0.0254 & 0.4668 & 0.6131  & 0.93 \\
			
			RoSD~\cite{xiao2025ICLRsafediffuser}    & 5.42\%  & 0.0303 & 0.4747 & 0.0837 &  29.84 \\
			TVSD~\cite{xiao2025ICLRsafediffuser}    & 15.51\%  & 0.0241 & 0.3283 & 0.0311 &  27.62 \\
			ReSD~\cite{xiao2025ICLRsafediffuser}    & 9.38\%  & 0.0257 & {0.0629} & 0.1068 &  32.54 \\
			
			CoB~\cite{Mizuta2024IROScoBLDiffusion} & 9.59\%  & {{0.0248}} & {0.5548} & {0.6561} &  {1.29}\\
			CoBL~\cite{Mizuta2024IROScoBLDiffusion} & 2.40\% & $\mathrm{>10^{16}}$ & {0.5102} & {0.7294} &  {4.32}\\
			\hline
			Diffuser~\cite{Janner2022ICMLplanning}    & 10.58\% & \textbf{0.0224} & \textbf{\textcolor{gray}{0.0522}} & 0.2858  & \textbf{\textcolor{gray}{0.31}}\\
			
			SafeDiffus(Ours)$\!\!\!$ & \textbf{100.00\%}  & \textbf{\textcolor{gray}{0.0233}} & {0.0985} & {0.0589} &  {0.54} \\
			\hline
			FM~\cite{lipman2023ICLRflow}    & 10.96\%  & 0.0265 & \textbf{0.0040} & \textbf{0.0001} &  \textbf{0.26}\\
			
			SafeFlow(Ours)$\!\!\!$ & \textbf{100.00\%}  & {0.0291} & 0.1150 &  \textbf{\textcolor{gray}{0.0062}}  & 0.38 \\
			\hline
		\end{tabular}
		\vspace{-0.3cm}
	\end{table}
	
	The simulation results of the robot navigation task in \cref{fig_car12} evaluate the performance of 11 trajectory generation approaches, namely SafeDiffuser (Ours), SafeFlow (Ours), Diffuser, FM, CG, TRUNC, CoB, CoBL, RoSD, ReSD, and TVSD.
	Each approach is evaluated based on 15 samples each, depicted in a 2D X-Y plane with blue regions indicating obstacles. 
	Training Trajectories are included to illustrate the variability in paths and, more importantly, to highlight that the training data lacks any consideration for obstacle avoidance. 
	Our proposed methods, SafeDiffuser and SafeFlow, achieved a perfect 100\% success rate in evading obstacles, showcasing outstanding reliability and safety. 
	SafeFlow, in particular, generated highly consistent and smooth trajectories that effectively avoided obstacles. 
	SafeDiffus also ensured obstacle avoidance, though it exhibited some outliers and less smooth trajectories compared to SafeFlow. 
	This phenomenon is also observed in the vanilla Diffuser, which does not account for obstacles, emphasizing the advantage of FM models.
	In contrast, methods such as CoB, ReSD, and RoSD faced significant challenges. 
	While these approaches attempted to avoid obstacles, their trajectories displayed abrupt jumps, causing them to penetrate obstacles and rendering the paths entirely unusable. 
	CoBL, on the other hand, represented an opposite extreme, becoming trapped between obstacles in an effort to avoid them, thus failing to reach the intended goal. 
	TVSD, TRUNC and CG managed to produce a few successful trajectories but were inconsistent in their ability to reliably avoid obstacles.
	
\begin{figure}[t]
		\centering
		\begin{subfigure}[b]{0.15\textwidth}
			\includegraphics[width=\linewidth]{fig_nav/Dataset.png}
			\caption{Dataset}
		\end{subfigure}
		\hfill
		\begin{subfigure}[b]{0.15\textwidth}
			\includegraphics[width=\linewidth]{fig_nav/SafeDiff.png}
			\caption{SafeDiff (Our)}
		\end{subfigure}
		\hfill
		\begin{subfigure}[b]{0.15\textwidth}
			\includegraphics[width=\linewidth]{fig_nav/SafeFlow.png}
			\caption{SafeFlow (Our)}
		\end{subfigure}
		
		\begin{subfigure}[b]{0.15\textwidth}
			\includegraphics[width=\linewidth]{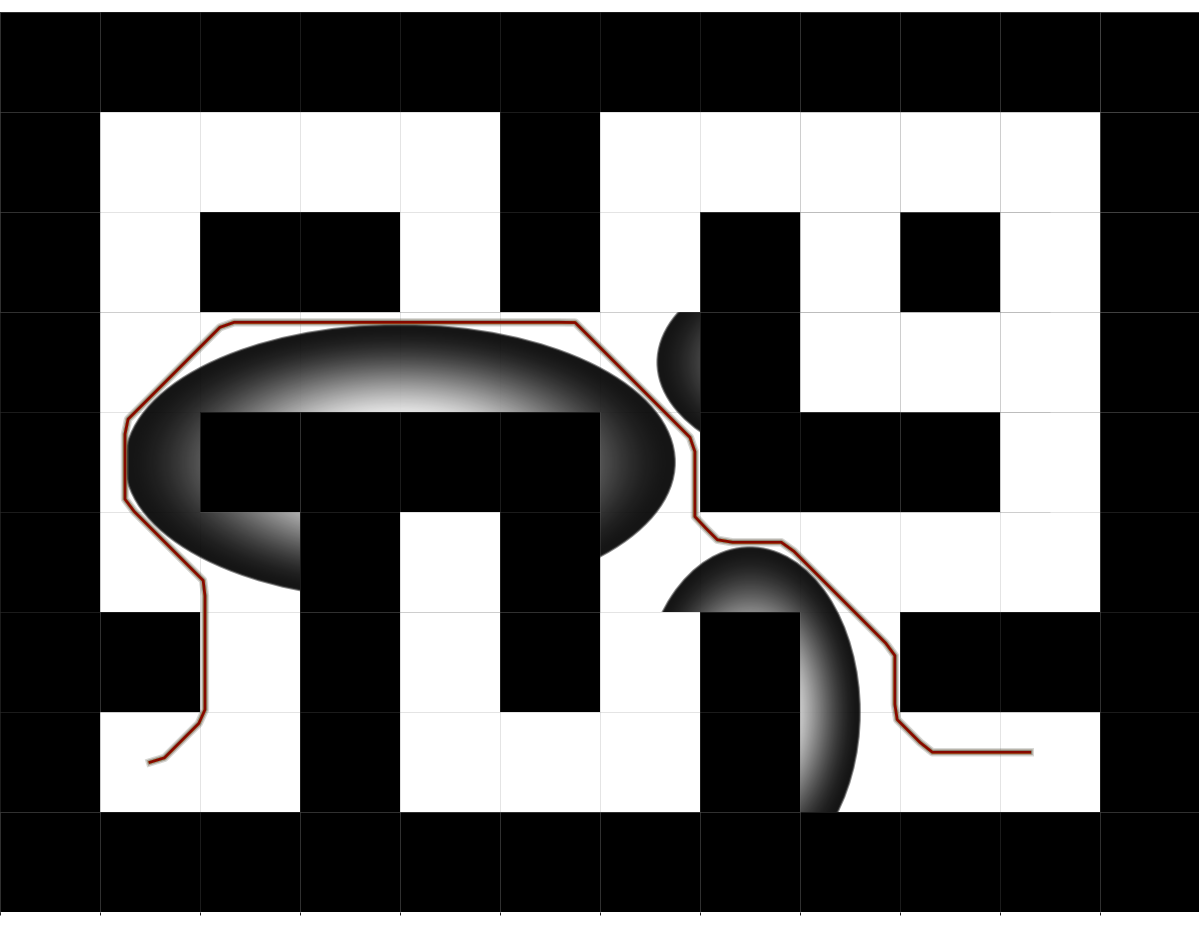}
			\caption{A*}
		\end{subfigure}
		\hfill
		\begin{subfigure}[b]{0.15\textwidth}
			\includegraphics[width=\linewidth]{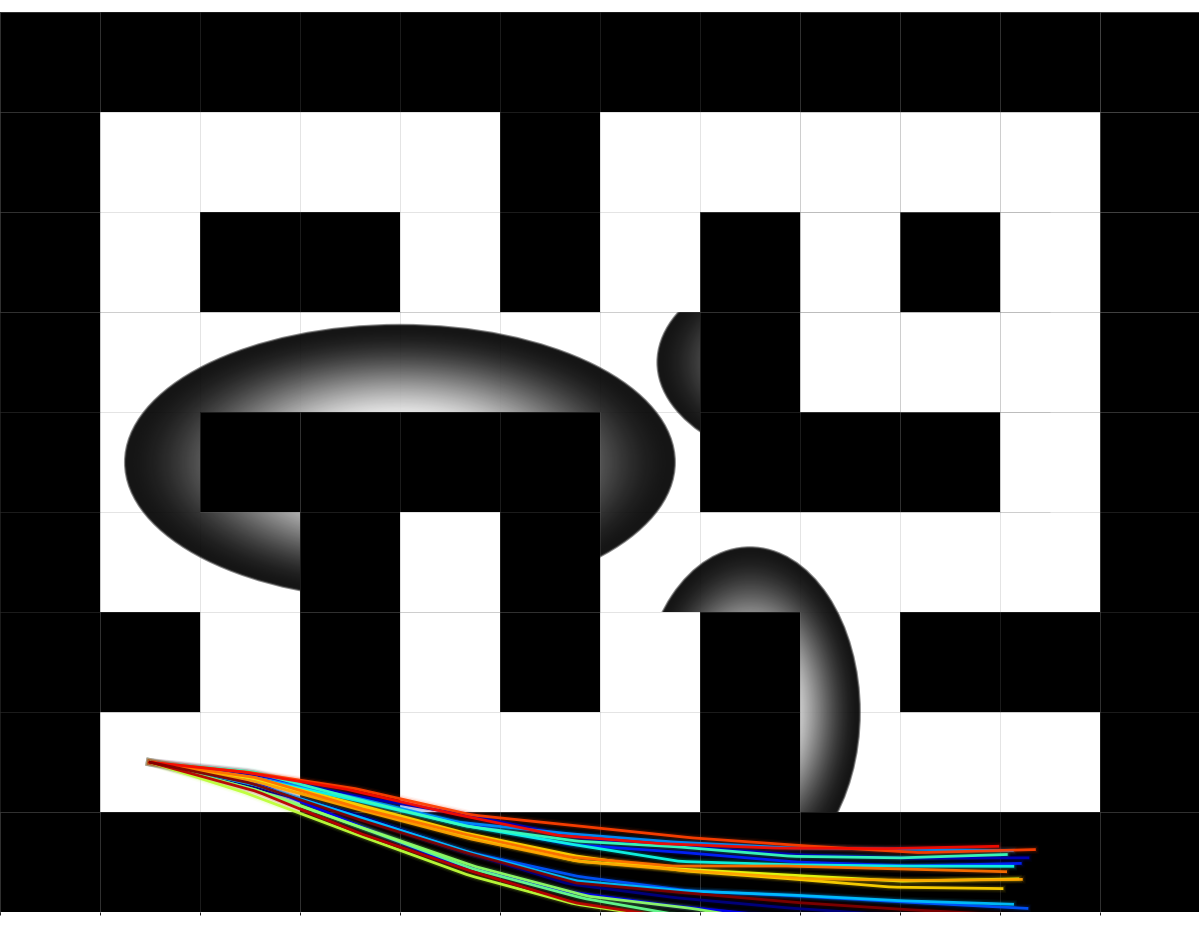}
			\caption{MPPI}
		\end{subfigure}
		\hfill
		\begin{subfigure}[b]{0.15\textwidth}
			\includegraphics[width=\linewidth]{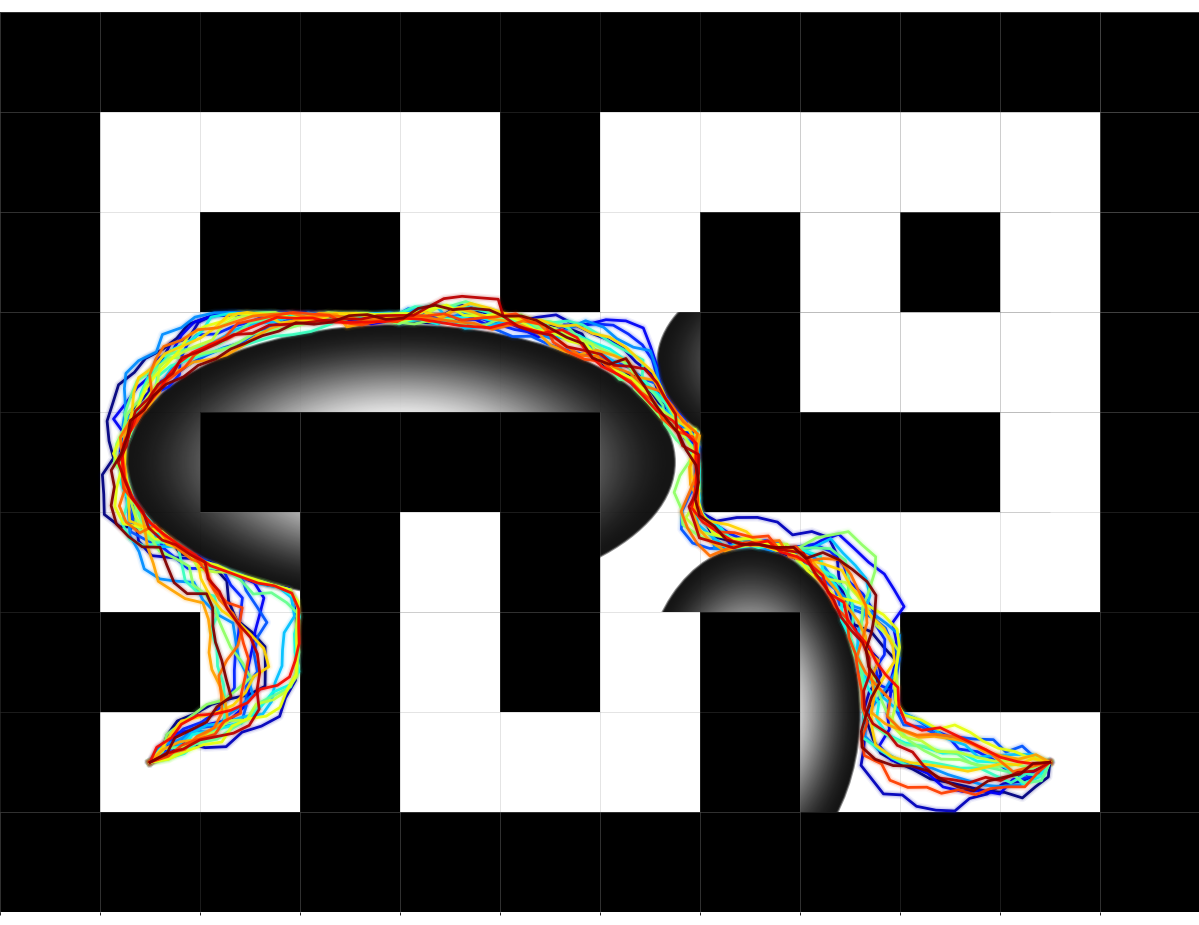}
			\caption{RRT*}
		\end{subfigure}
		\vspace{-0.3cm}
		\caption{Trajectories for robot navigation using various tradition motion planning methods.}
		\label{figure_traditional_method}
		\vspace{-0.5cm}
	\end{figure}

	In addition, we also compare our proposed SafeFlow and SafeDiffuser with some traditional motion planning approaches, including A*~\cite{zeng2009finding}, MPPI~\cite{kim2022smooth} and RRT*~\cite{noreen2016optimal}.
	The results are shown in \cref{figure_traditional_method} and summarized in \cref{tab_manipulation_tradition}.
	While MPPI achieves a low safety rate, most trajectories generated by A* and RRT* are safe, due to their consideration of obstacles in the algorithm.
	However, they are more time consuming compared to our proposed SafeFlow and SafeDiffuser.
	
	\begin{table}[t]
		\centering
		\caption{Safe planning for robot manipulation comparisons to traditional path planning approaches.}
		\label{tab_manipulation_tradition}
		\begin{tabular}{l c c c c c c }
			\hline
			\textbf{Method} & $\!\!$\textbf{Safety$\uparrow\!$ } & $\!\!$\textbf{Time$(\!s\!)\!\downarrow\!$} & $\!\!$\textbf{SD$\downarrow\!$} & 
			$\!\!$\textbf{ED$\downarrow\!$} & $\!\!$\textbf{CS$\downarrow\!$ } & $\!\!$\textbf{AS$\downarrow\!$} \\
			\hline 
			A*~\cite{zeng2009finding}    &\textbf{100.0\%} & 2.847 & \textbf{0.000} & \textbf{0.000} & \textbf{0.120} &\textbf{\textcolor{gray}{0.021}} \\
			MPPI~\cite{kim2022smooth}    &~0.00\% & 100+ & \textbf{0.000} & --- & --- & --- \\
			RRT*~\cite{noreen2016optimal}    & 98.62\%  & 6.806 & \textbf{0.000} & \textbf{0.000} & 0.324 & 0.073 \\
			\hline 
			SafeDiff $\!\!\!$ & \textbf{100.0\%}& \textbf{\textcolor{gray}{0.54}}  & \textbf{\textcolor{gray}{0.008}} & 0.013 & \textbf{0.099} & 0.059 \\
			SafeFlow $\!\!\!$ & \textbf{100.0\%} &\textbf{{0.38}} & \textbf{\textcolor{gray}{0.008}} & \textbf{\textcolor{gray}{0.007}}  & \textbf{\textcolor{gray}{0.115}}& \textbf{0.006}  \\
			\hline
		\end{tabular}
	\end{table}
	
	\subsection{Robot Manipulation}
	\label{subsec_experiment}
	
	For more complicated robot manipulation task, we consider the motion planning for a Franka Research 3 with $7$ degrees of freedom, which moves from initial joint configuration $\bm{q}_{0}^{*} = \bm{0}_{7 \times 1}$ to the target $\bm{q}_{target}^{*}$ with
	\begin{align} 
		\bm{q}_{target}^{*} = [2.0,\;1.5,\;2.0,\;-2.0,\;2.0,\;3.0,\;2.0]^{\top}.
	\end{align} 
	The training data generation process is shown in Appendix B.
	
	While no obstacles are considered in data generation phase, two spherical obstacles are added in the motion generation part, where each obstacle with center \(\bm{c}_j\in\mathbb{R}^3\) and radius \(r_j \in \mathbb{R}_{>0}\), $j=1,2$. We define the barrier function as
	\begin{align}
		h_j(\bm{E}_k\bm{\mathcal{T}}) = \|\bm{x}_e(\bm{s}^i) - \bm{c}_j\| -  \beta r_j,
	\end{align}
	for all $i = 0, 1,\dots, H$ and $j = 1,2$, where \(\bm{s}^i\in\mathbb{R}^7\) are the joint angles of the robot arm, \(\bm{x}_e(\bm{s}^i)$ is the end-effector position in Euclidean space computed via MuJoCo forward kinematics corresponding to the joint angles $\bm{s}^i$, and \(\beta\) is an expansion factor is set as 0.25.
	The obstacle positions are 
	\begin{align}
		\bm{c}_1 = [-0.62, 0.30, 0.50]^{\top}, &&
		\bm{c}_2 = [-0.05, 0.78, 0.55]^{\top}
	\end{align}
	with radius $r_1=0.125$ and $r_2=0.25$ m, respectively. 
	
\begin{figure*}[t]
		\centering
		\begin{subfigure}[b]{0.15\textwidth}
			\includegraphics[width=\linewidth]{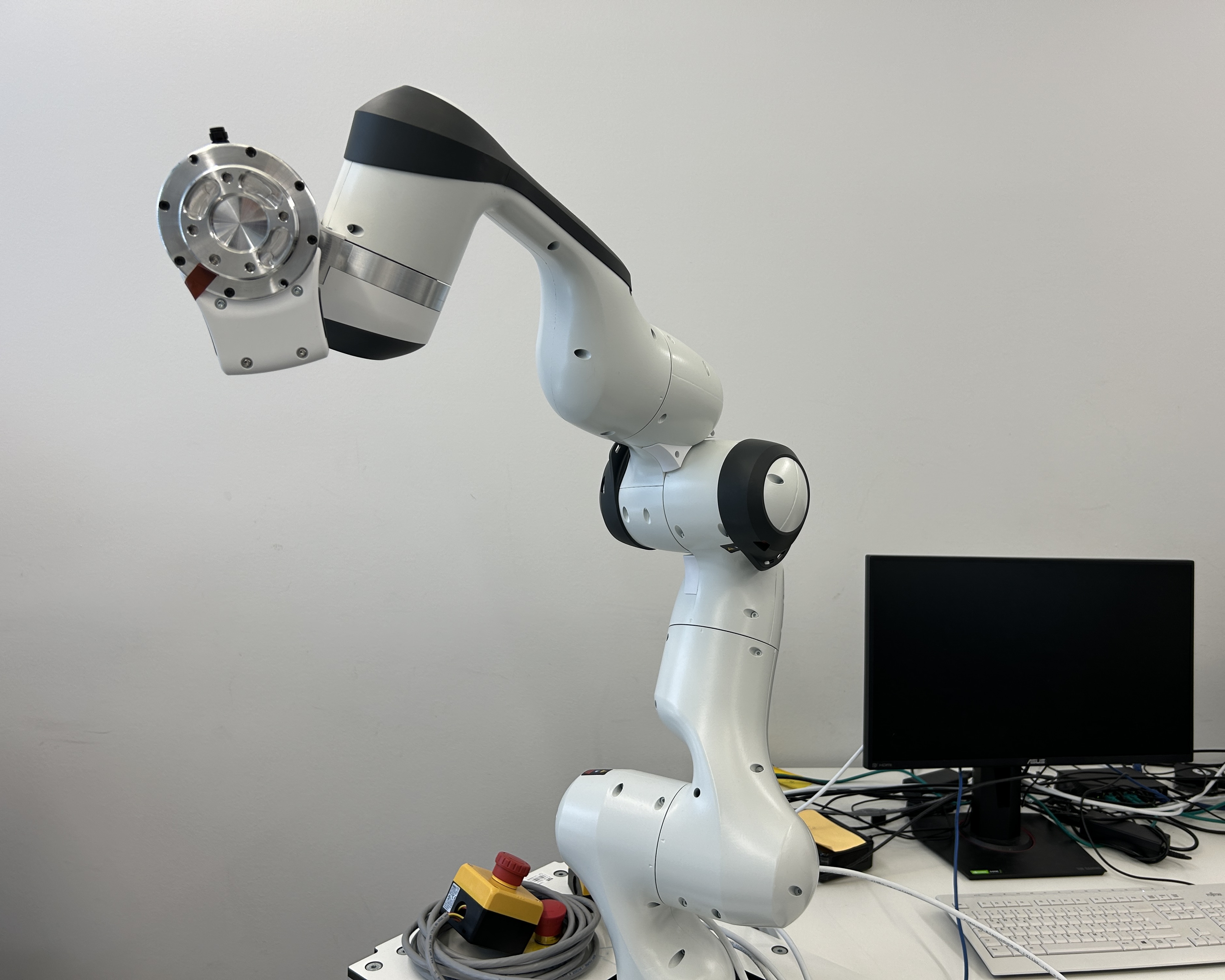}
			\caption{Franka}
			\label{fig:franka}
		\end{subfigure}
		\hfill
		\begin{subfigure}[b]{0.15\textwidth}
			\includegraphics[width=\linewidth]{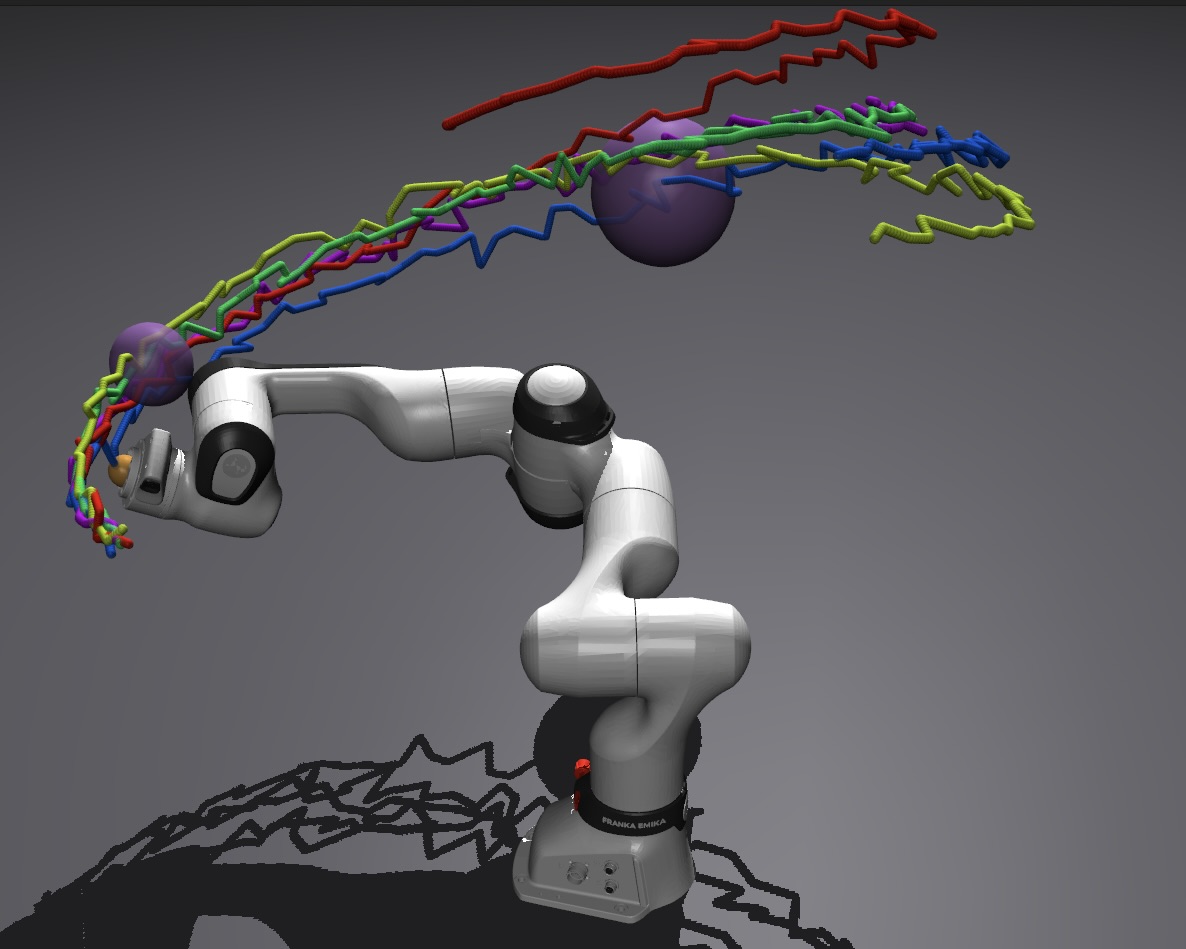}
			\caption{Diffuser}
		\end{subfigure}
		\hfill
		\begin{subfigure}[b]{0.15\textwidth}
			\includegraphics[width=\linewidth]{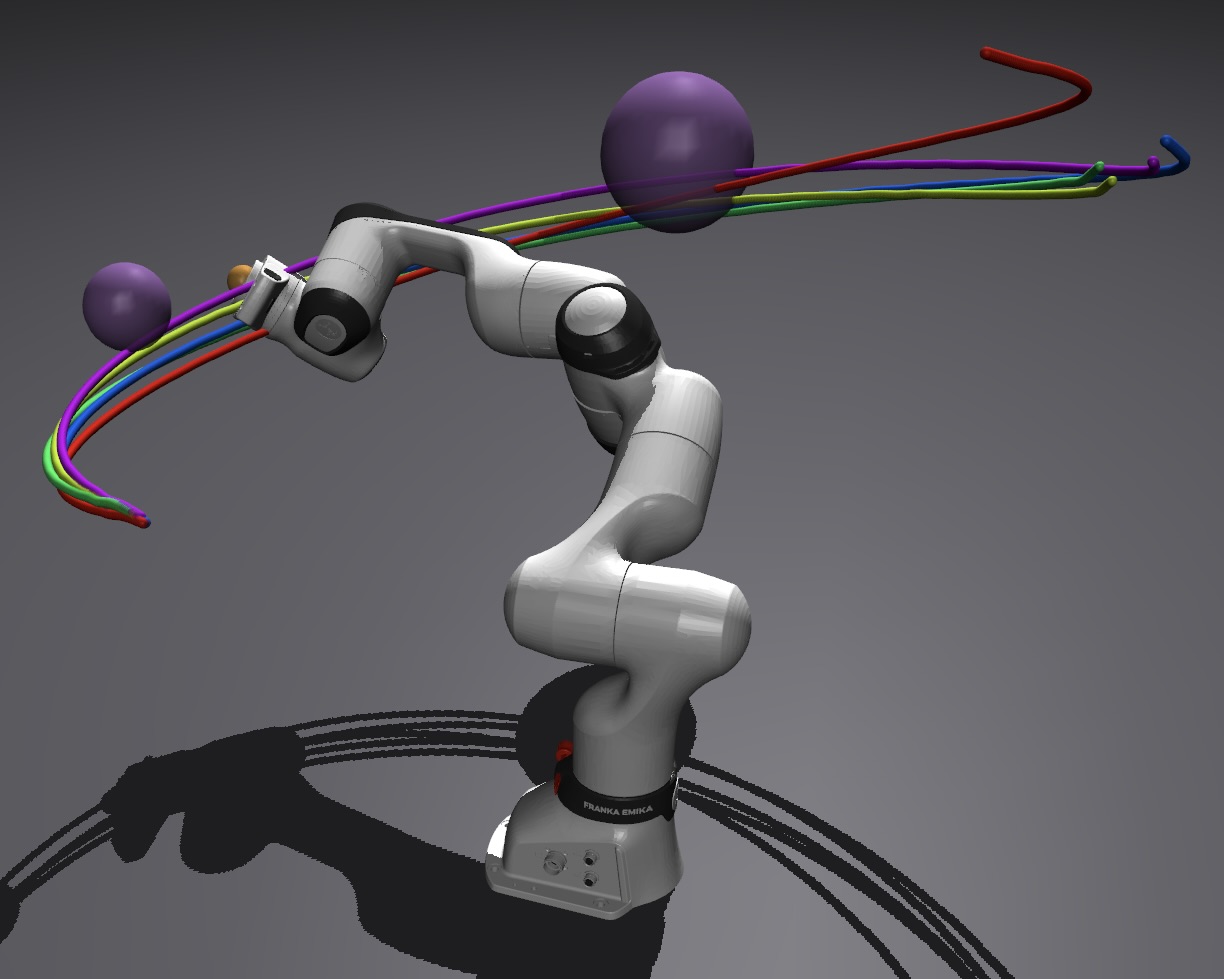}
			\caption{FM}
		\end{subfigure}
		\hfill
		\begin{subfigure}[b]{0.15\textwidth}
			\includegraphics[width=\linewidth]{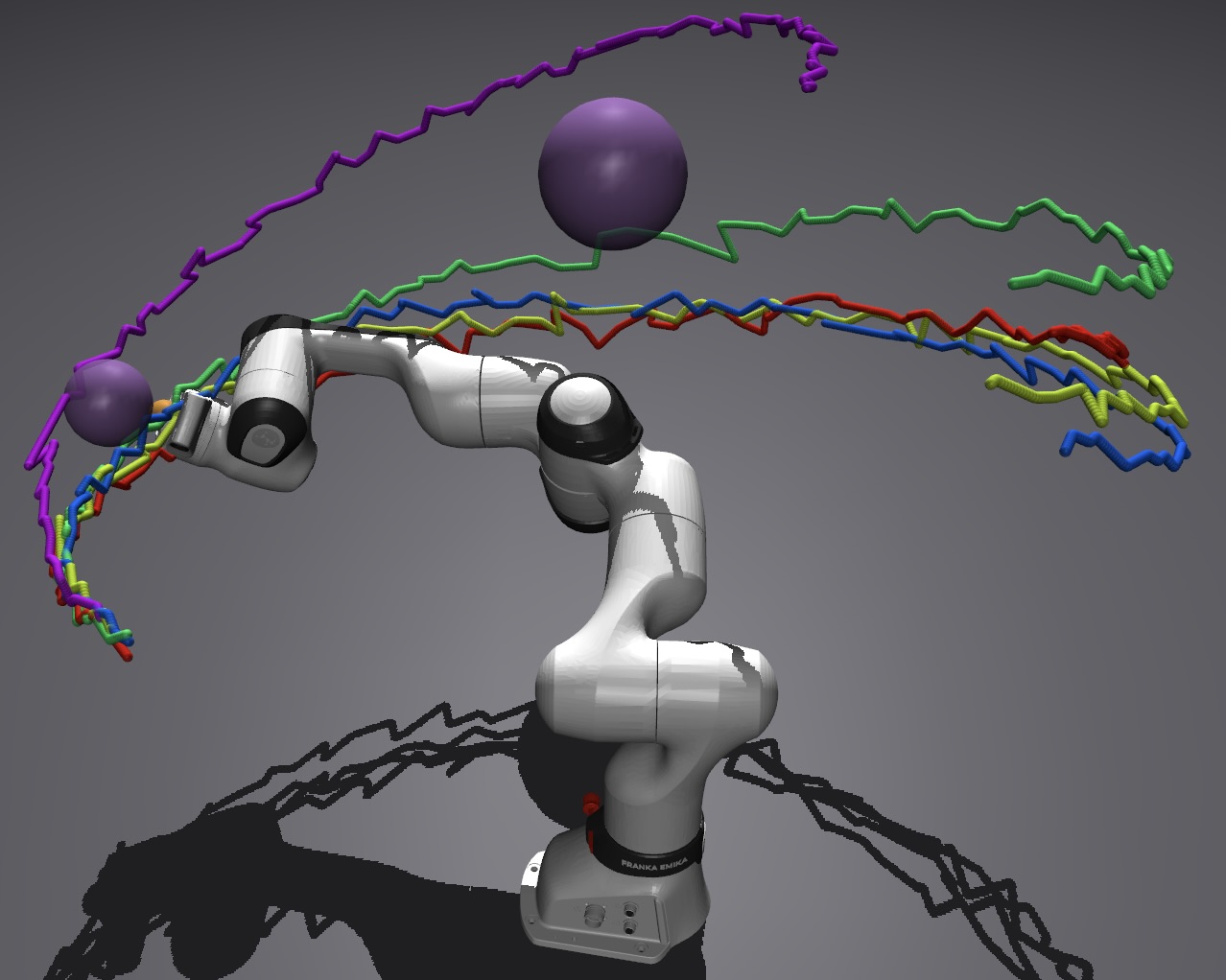}
			\caption{CG}
		\end{subfigure}
		\hfill
		\begin{subfigure}[b]{0.15\textwidth}
			\includegraphics[width=\linewidth]{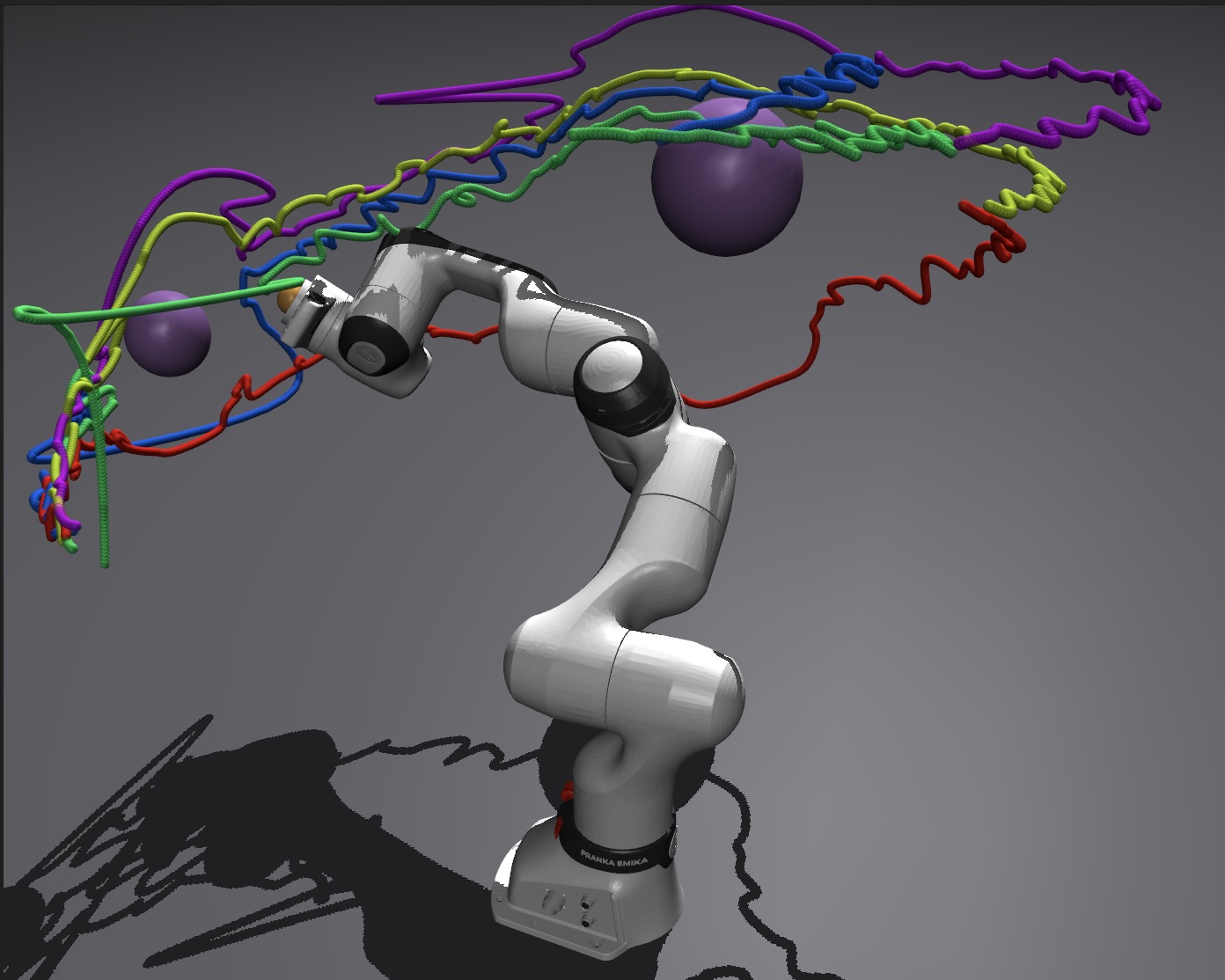}
			\caption{SafeDiffus (Ours)}
		\end{subfigure}
		\hfill
		\begin{subfigure}[b]{0.15\textwidth}
			\includegraphics[width=\linewidth]{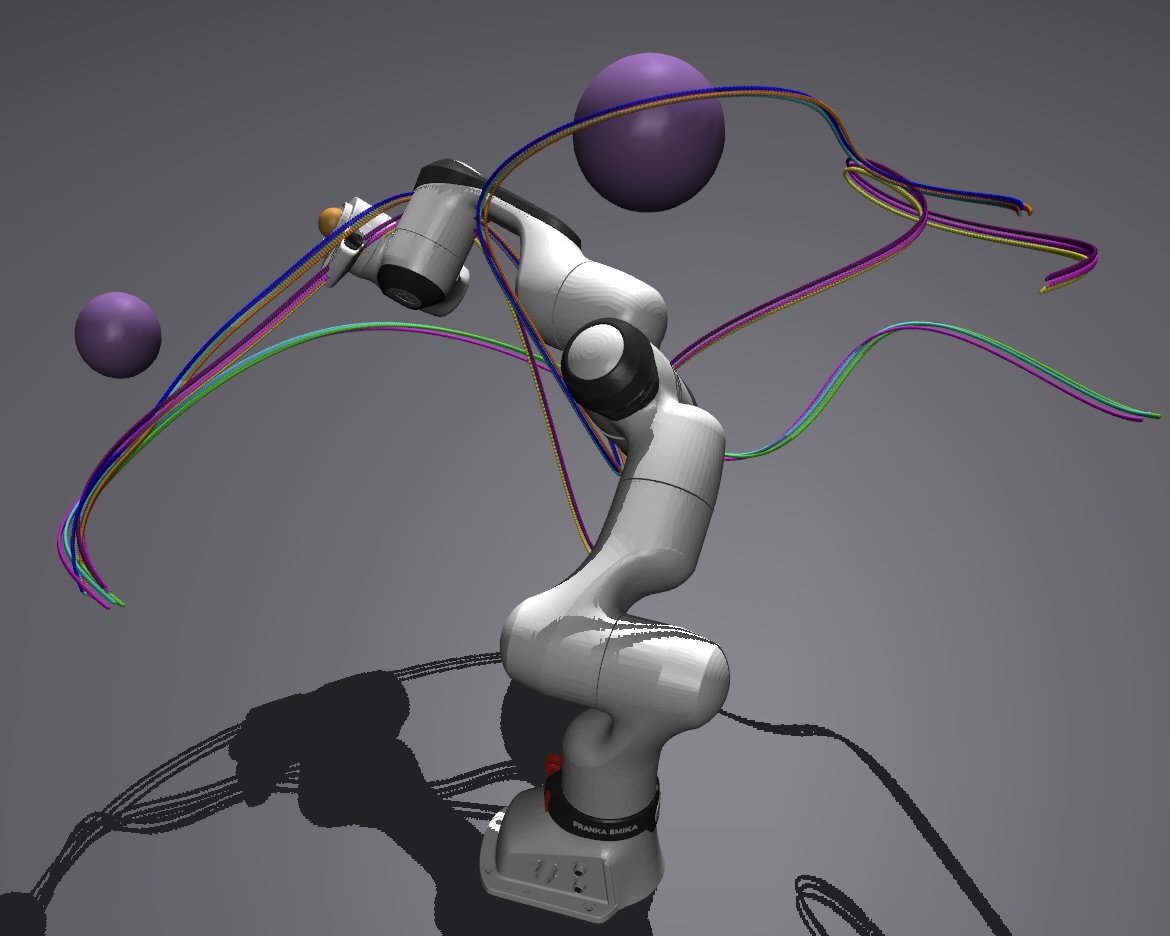}
			\caption{SafeFlow (Our)}
		\end{subfigure}
		
		\begin{subfigure}[b]{0.15\textwidth}
			\includegraphics[width=\linewidth]{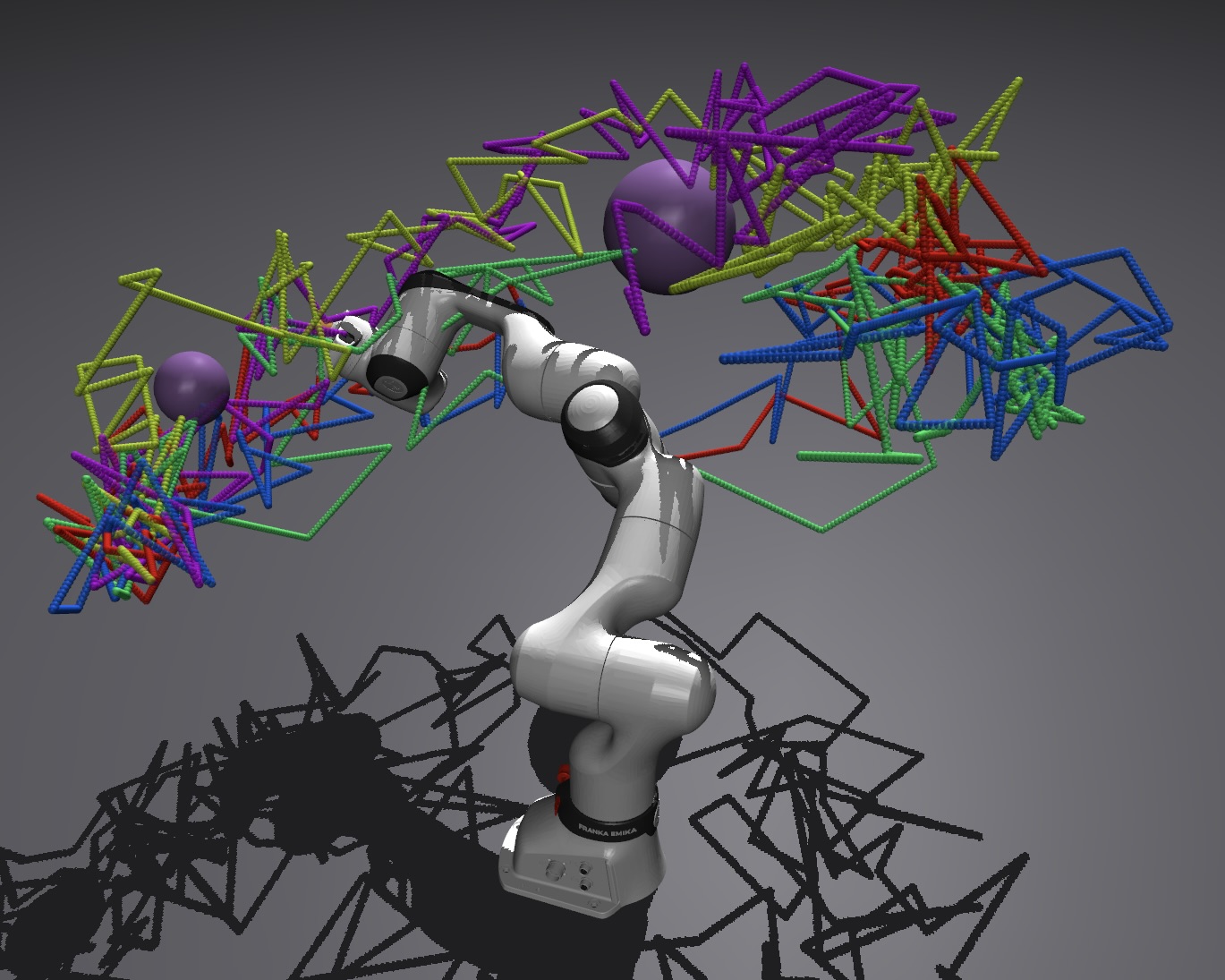}
			\caption{TRUNC}
		\end{subfigure}
		\hfill
		\begin{subfigure}[b]{0.15\textwidth}
			\includegraphics[width=\linewidth]{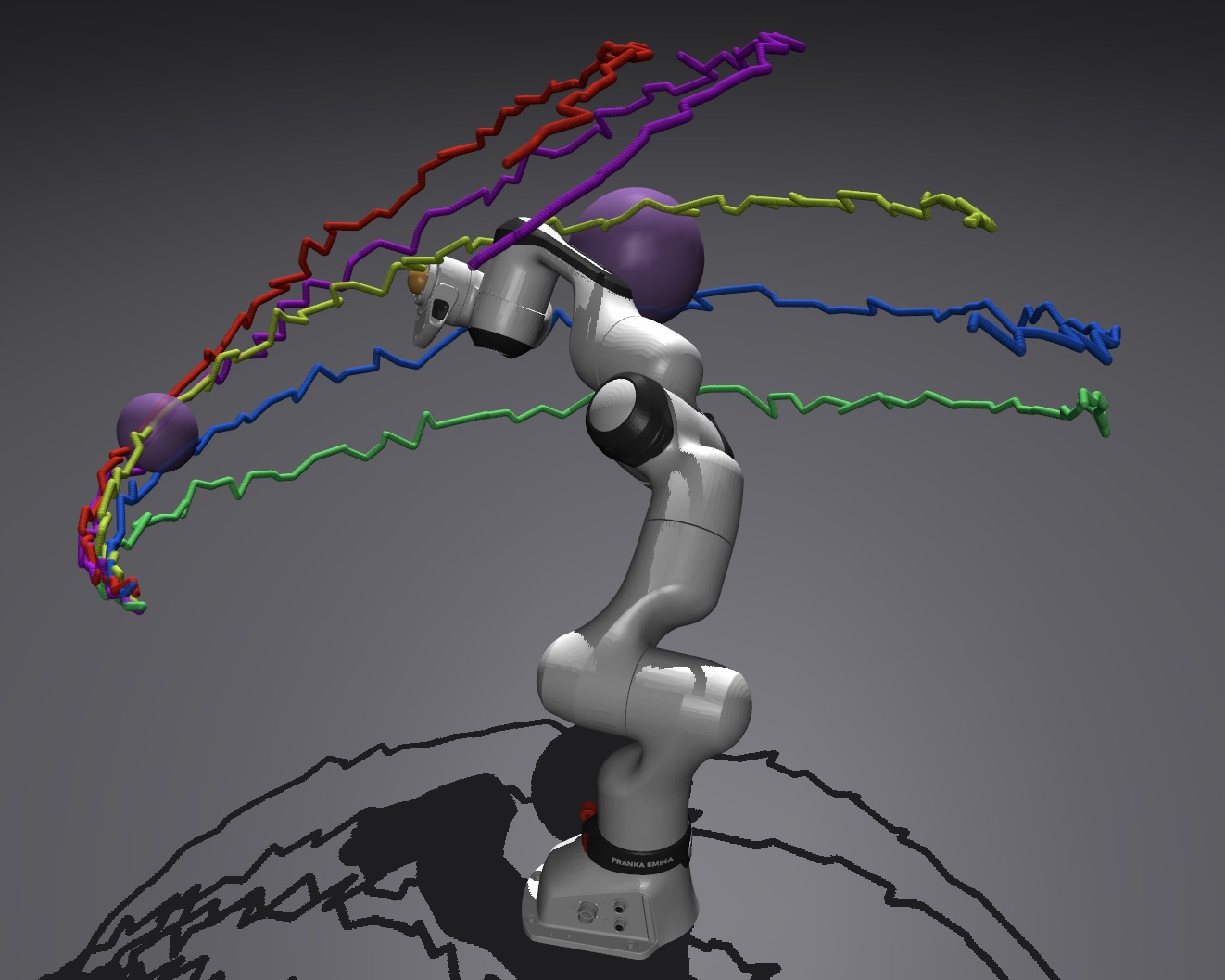}
			\caption{CoB}
		\end{subfigure}
		\hfill
		\begin{subfigure}[b]{0.15\textwidth}
			\includegraphics[width=\linewidth]{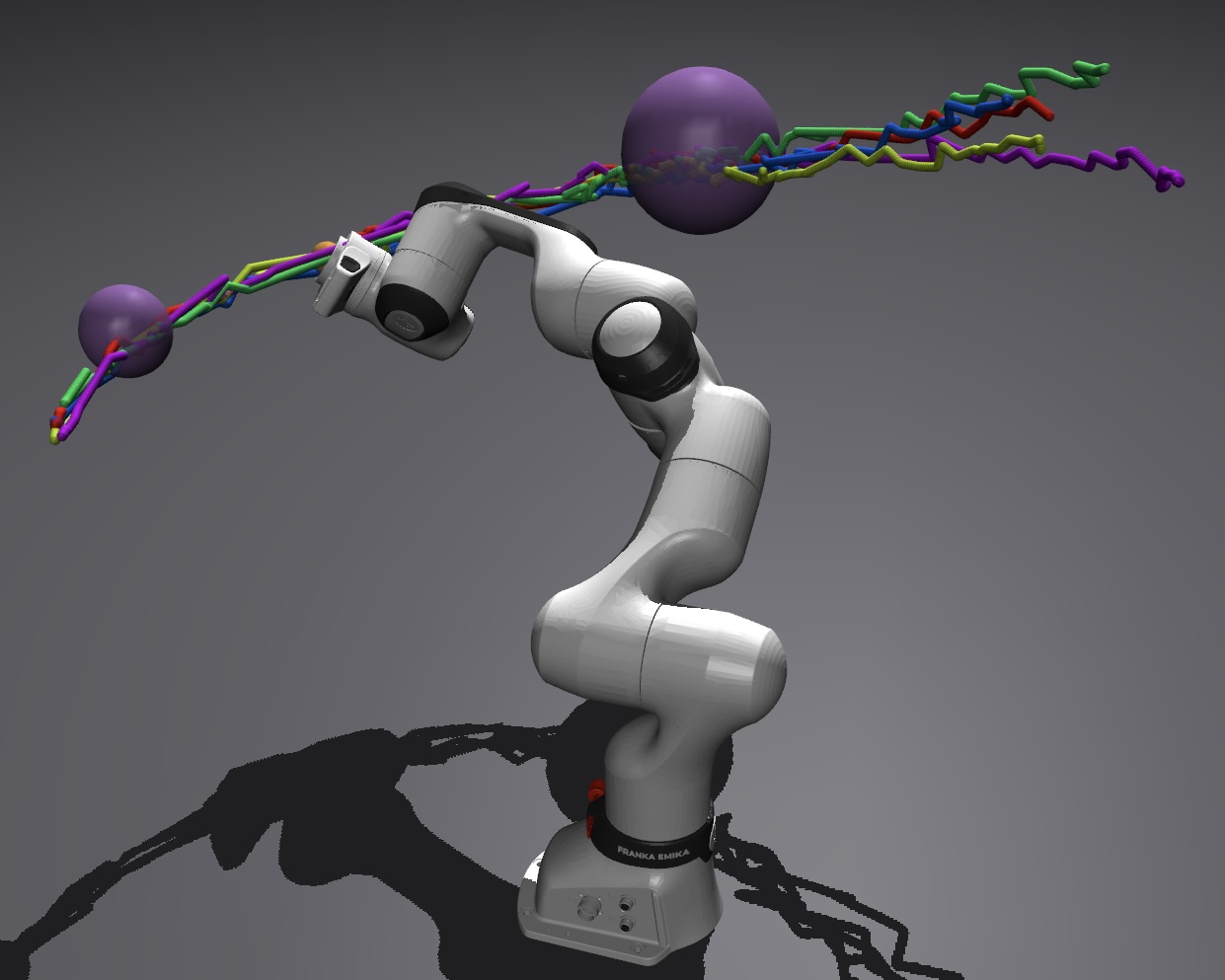}
			\caption{CoBL}
		\end{subfigure}
		\hfill
		\begin{subfigure}[b]{0.15\textwidth}
			\includegraphics[width=\linewidth]{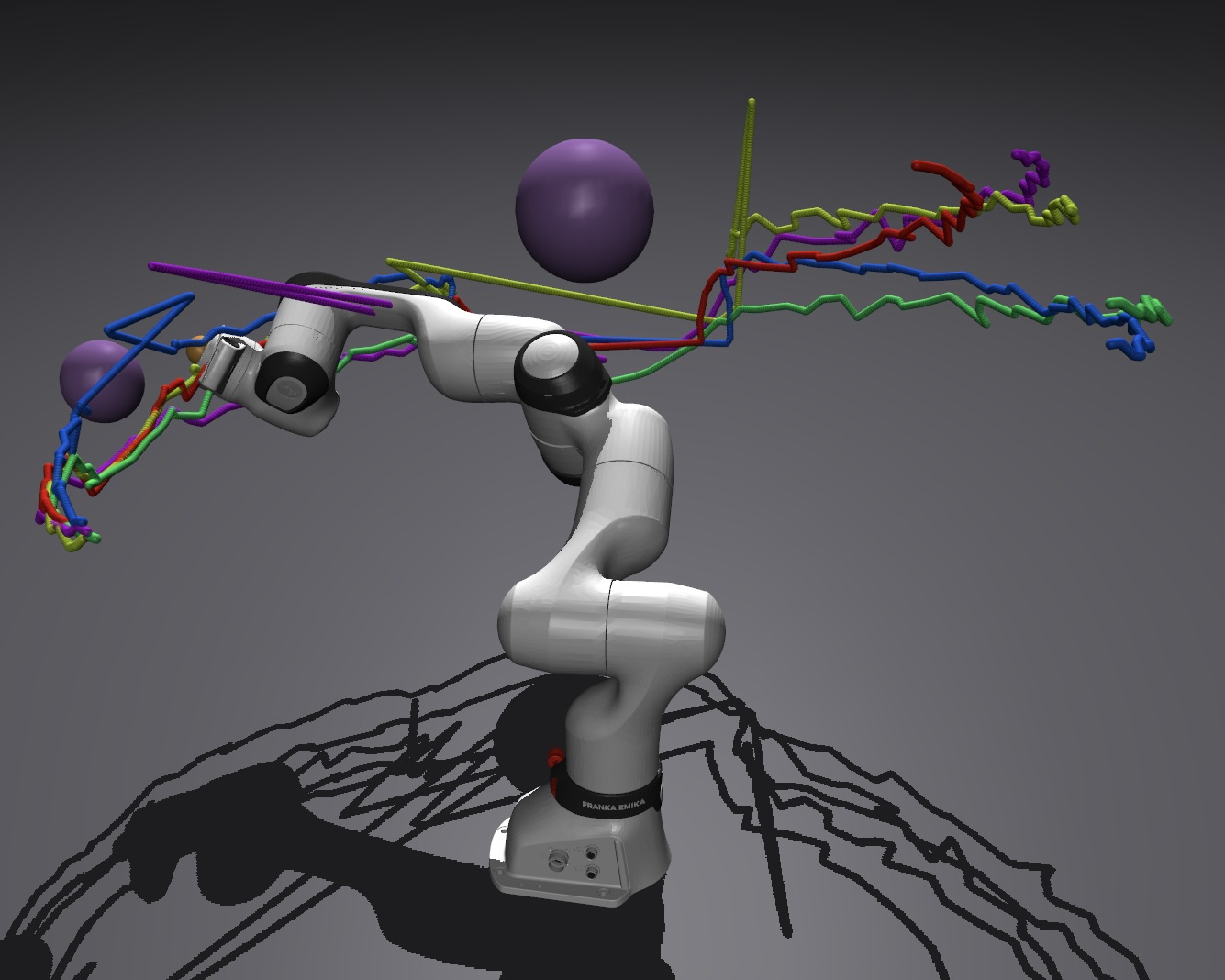}
			\caption{RoSD}
		\end{subfigure}
		\hfill
		\begin{subfigure}[b]{0.15\linewidth}
			\includegraphics[width=\linewidth]{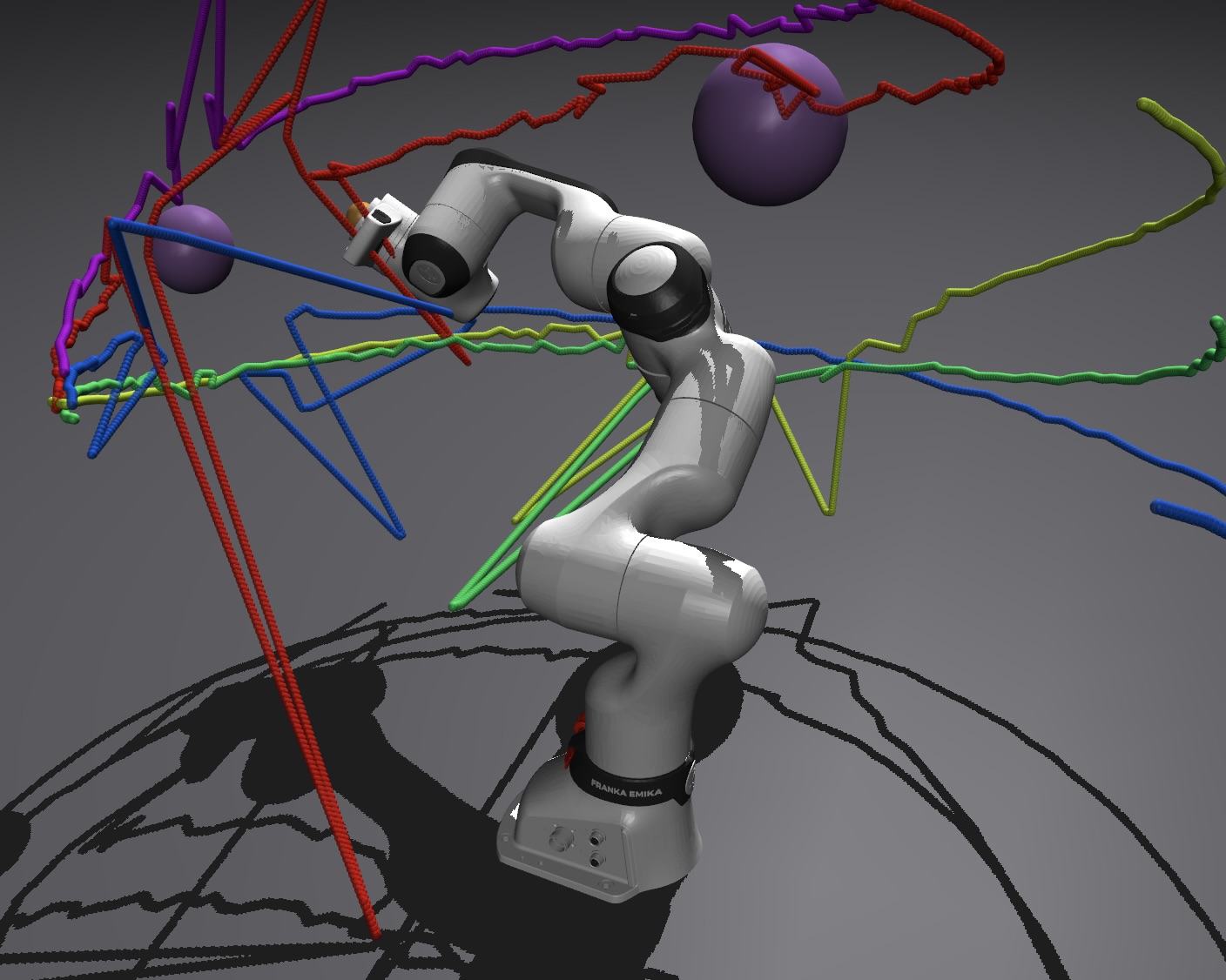}
			\caption{ReSD}
		\end{subfigure}
		\hfill
		\begin{subfigure}[b]{0.15\linewidth}
			\includegraphics[width=\linewidth]{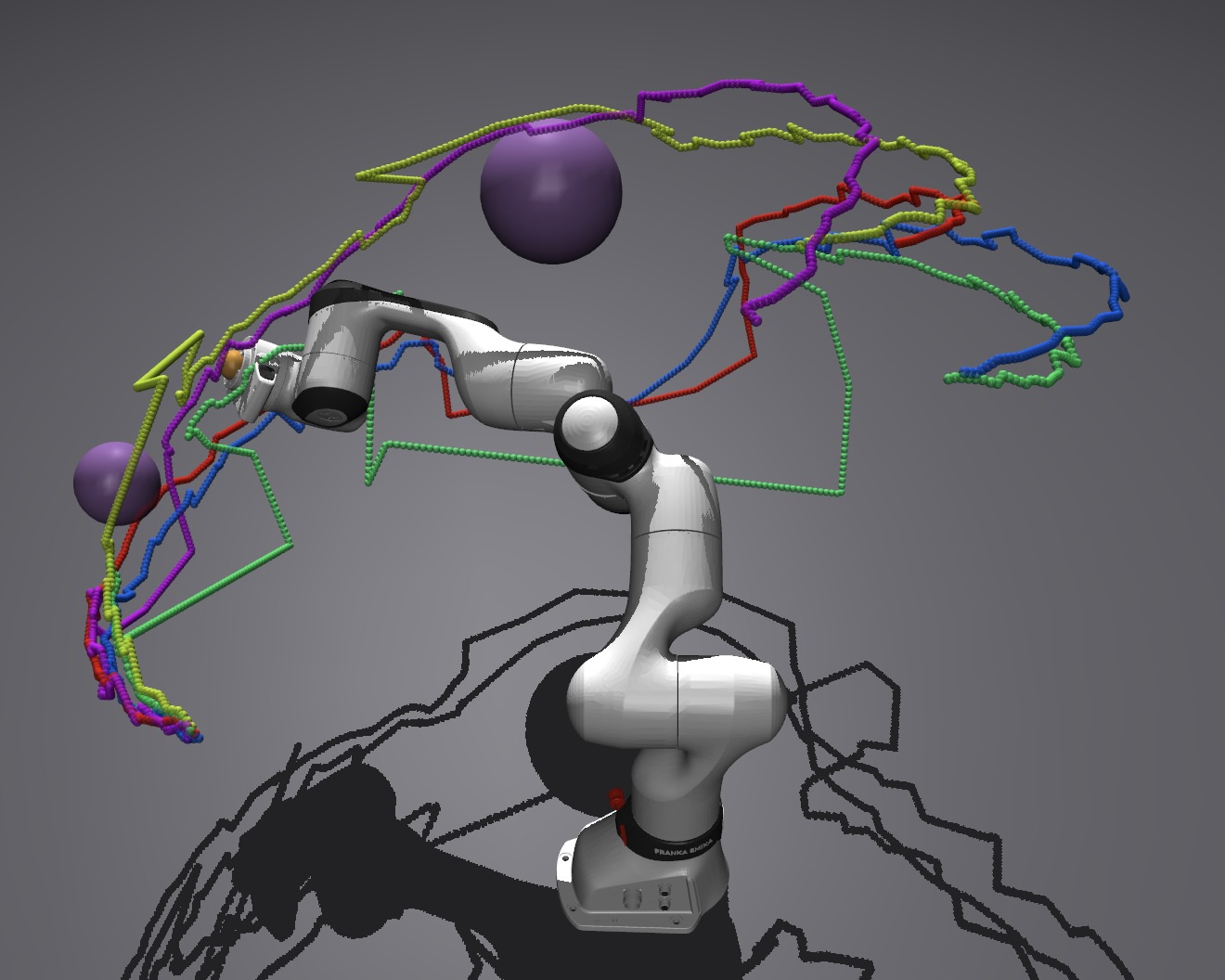}
			\caption{TVSD}
		\end{subfigure}
		\vspace{-0.3cm}
		\caption{End-effector trajectories for robot manipulation using various methods.}
		\vspace{-0.3cm}
		\label{fig_trajectories_4x3}
	\end{figure*}

	To demonstrate the effectiveness of the proposed SafeFlow and SafeDiffus methods, we evaluate the same metrics used in \cref{subsec_planar} for a robotic manipulation task using Franka Research 3 as shown in \cref{fig_trajectories_4x3}.
	Moreover, we also measured bias between the starting (end) state of the generalized trajectory and the desired start (end) point, denoted as starting distance (SD) and end distance (ED).
	Specifically, the SD score $S_D \in \mathbb{R}_{0,+}$ and ED score $E_D \in \mathbb{R}_{0,+}$ are 
	\begin{align}
		&S_D = \| \bm{E}_0 \bm{\mathcal{T}} - [(\bm{q}_{0}^{*})^{\top}, \bm{0}_{14 \times 1}^{\top}]^{\top} \|, \\
		&E_D = \| \bm{E}_H \bm{\mathcal{T}} - [(\bm{q}_{target}^{*})^{\top}, \bm{0}_{14 \times 1}^{\top}]^{\top} \|.
	\end{align}
	The objective is to move the robot arm from a lower to an upper position without collision with obstacles, which are unseen in the training datasets.
	The experiment is repeated $1000$ times for each method, and the performance of each method is evaluated based on the metrics mentioned in \cref{subsection_general_simulation_setting} for the motion trajectory of the end-effector, which is summarized in \cref{tab_manipulation}.
	It is obvious to see that the proposed SafeFlow demonstrates clear superiority in KL, CS, and inference time (except vanilla diffusion and flow models), especially in the safety aspect with 100\% safety guarantees, providing a strong affirmative to \textbf{Q1}. 
	It is noted that some baseline methods (Diffuser, FM, and ReSD) have low CS or AS due to the generation of overly simplified trajectories, lacking the effective response to obstacles and inducing insecurity.
	Therefore, SafeFlow demonstrates better performance compared to the SOTA methods~\cite{xiao2025ICLRsafediffuser, Mizuta2024IROScoBLDiffusion, Dhariwal2021NeurIPSDiffusion, brockman2016openaigym} that integrate safe constraints, which answers \textbf{Q3}. 
	Similar as in \cref{subsec_planar}, there is a slight degradation in performance compared to vanilla approaches (Diffuser and FM), which is expected, as the FMBFs influence the generation process during inference to enforce safety. 
	
	\begin{table}[t]
		\centering
		\caption{Safe planning for robot manipulation comparisons.}
		\label{tab_manipulation}
		\begin{tabular}{l c c c c c c }
			\hline
			\textbf{Method} & $\!\!$\textbf{Safety$\uparrow\!$ }  & $\!\!$\textbf{Time$(\!s\!)\!\downarrow\!$}
			&$\!\!$\textbf{SD$\downarrow\!$} & 
			$\!\!$\textbf{ED$\downarrow\!$} & $\!\!$\textbf{CS$\downarrow\!$ } & $\!\!$\textbf{AS$\downarrow\!$}  \\
			\hline

			CG    & 81.90\%  & 4.1379 & 0.0109 &  0.0177 & 0.1820 & 0.0555\\
			
			TRUNC    & 0.20\%  & 3.4185 & 6.7571 &  0.1773 & 0.4241 & 2.1514\\
			
			RoSD    & 48.50\%  & 93.0668 & 0.0129 & 0.0110 & 0.1716 &   0.0482\\
			TVSD    & 67.60\% &  82.7284 & 0.0141 & 0.0113 & 0.1713 &   0.0514 \\
			ReSD   & 68.00\%  & 62.3659 & 0.0143 &  0.0113 & 0.1707 &   0.0493\\
			CoB & 71.80\%  & 4.1457 & 0.0188 & 0.0054 &  0.2066 & 0.0625\\
			CoBL &~0.00\%  & 4.5179 & 1.5545 & 0.8182 &  0.5146 & 13.5480\\
			\hline
			Diffuser   &~0.00\%  & 1.4512 & \textbf{0.0031} &  0.0040 & \textbf{0.1636} & 0.0351\\
			
			SafeDiff $\!\!\!$ & \textbf{100.0\%}& 2.0711  & 0.0078 & 0.0130 & 0.1617 & 0.0459 \\
			\hline
			FM   &~0.00\%  & \textbf{{0.5578}} & \textbf{\textcolor{gray}{0.0042}} & \textbf{0.0021}& 0.1639 & \textbf{0.0339}\\
			
			SafeFlow $\!\!\!$ & \textbf{100.0\%} &0.7540 & 0.0079 & \textbf{\textcolor{gray}{0.0074}}  & \textbf{\textcolor{gray}{0.1548}}& \textbf{\textcolor{gray}{0.0398}}  \\
			\hline
		\end{tabular}
		\vspace{-0.3cm}
	\end{table}
	
	The simulation results presented in \cref{fig_trajectories_4x3} illustrate the end-effector trajectories of all methods, using 15 samples each, with a robotic arm (Franka Research 3) in a 3D environment where a purple sphere represents the target. 
	Consistent with the robot navigation findings, our proposed methods, SafeDiffuser and SafeFlow, demonstrate exceptional obstacle avoidance capabilities. 
	SafeFlow generates smooth and highly consistent trajectories that reliably reach the target while avoiding obstacles, reinforcing its superiority in precision and safety as highlighted in the robot navigation section. 
	SafeDiffus also successfully avoids obstacles, though its trajectories exhibit less smoothness compared to SafeFlow. 
	In contrast, baseline methods such as Diffuser, FM, and CG produce trajectories with varying degrees of success, but they often show erratic paths or occasional collisions with obstacles. 
	Methods like TRUNC, RoSD, and ReSD generate highly unstable trajectories, frequently intersecting obstacles, rendering them impractical.
	CG, CoB and CoBL achieve limited success but lack consistency, while TVSD shows moderate improvement yet still falls short of our methods’ reliability. 
	These results further validate the robustness of SafeFlow and SafeDiffuser for safety-critical robotic applications.
	
	\section{Conclusion}
	\label{sec_conclusion}
	In this paper, we addressed the critical challenge of enabling robots to generate safe and efficient trajectories in dynamic and unseen obstacles, a fundamental requirement for autonomous systems in real-world applications. 
	Existing generative methods, such as diffusion models, while capable of modeling complex trajectory distributions, often incur high computational costs and lack explicit safety guarantees, limiting their practical applications in safety-critical scenarios. 
	To address these shortcomings, we proposed SafeFlow, a novel motion planning framework that integrates flow matching with FMBFs inspired by CBFs.
	Our proposed SafeFlow ensures trajectory safety while maintaining computational efficiency.
	Specifically, SafeFlow leverages the deterministic and efficient nature of flow matching to transform samples from a simple prior distribution into goal-directed trajectories, embedding safety constraints directly into the learned flow field via FMBFs. 
	This approach guarantees that generated trajectories remain within safe regions throughout the planning horizon, even in the presence of unseen obstacles or dynamic constraints. 
	Our experimental evaluations, including planar robot navigation and 7-DoF robot manipulation, demonstrated the superior performance of SafeFlow over SOTA planners. 
	These advantages make it suitable for real-time safety-critical applications.
	
	\bibliographystyle{IEEEtran}
	\bibliography{ref}

\end{document}